\documentclass[10pt,twocolumn,letterpaper]{article}

\usepackage[pagenumbers]{cvpr} 

\newcommand{\algorithm}{MiDiffusion\xspace}

\newcommand{\myParagraph}[1]{{\bf #1.}\xspace}

\newcommand{\M}[1]{{\bm #1}}
\newcommand{\norm}[1]{\left\| #1 \right\|}
\newcommand{\E}{{\mathbb{E}}}   
\newcommand{\calN}{{\cal N}}
\newcommand{\calS}{{\cal S}}
\newcommand{\tran}{^{\mathsf{T}}}
\newcommand{\MI}{\M{I}}         
\newcommand{\Real}[1]{ { {\mathbb R}^{#1} } }
\newcommand{\Int}[1]{ { {\mathbb Z}^{#1} } }

\newcommand{\ContinuousVar}{\M{x}}
\newcommand{\GaussianMean}{\M{\mu}}
\newcommand{\GaussianCovar}{\M{\Sigma}}

\newcommand{\discreteVar}{z}
\newcommand{\DiscreteVar}{\M{z}}
\newcommand{\oneHotVec}[1]{\M{v}( #1 )}

\newcommand{\DiscreteTransMat}{\M{Q}}

\newcommand{\CondVar}{\M{y}}
\newcommand{\networkVar}{\theta}

\newcommand{\numObj}{N}
\newcommand{\numLabels}{C}
\newcommand{\posVar}{\M{t}}
\newcommand{\sizeVar}{\M{s}}
\newcommand{\angleVar}{\phi}

\definecolor{cvprblue}{rgb}{0.21,0.49,0.74}
\usepackage[pagebackref,breaklinks,colorlinks,allcolors=cvprblue]{hyperref}

\usepackage{comment}
\usepackage{bm}
\usepackage{amssymb}
\usepackage{amsmath}
\usepackage{multirow}
\usepackage{xspace}
\usepackage{graphicx}

\newcommand{\Update}[2]{{#1}}	

\def\paperID{13746} 
\def\confName{CVPR}
\def\confYear{2025}

\title{Mixed Diffusion for 3D Indoor Scene Synthesis} 

\author{
	\hspace{1cm} Siyi Hu$^1$ \and Diego Martin Arroyo$^2$ \and Stephanie Debats$^2$ \and Fabian Manhardt$^2$ \hspace{1cm}
	\and Luca Carlone$^1$ \and Federico Tombari$^{2,3}$ \\
	\and {} \vspace{-16pt} \\
	$^1$Massachusetts Institute of Technology \quad $^2$Google, Inc \quad $^3$Technische Universit\"{a}t M\"{u}nchen \\
	{\tt\small \{siyi, lcarlone\}@mit.edu} \and {\tt\small \{martinarroyo, sdebats, fabianmanhardt, tombari\}@google.com} \\
}

\begin{document}
\setlength{\belowcaptionskip}{-10pt}

\maketitle

\begin{abstract}

    Generating realistic 3D scenes is an area of growing interest in computer vision and robotics. However, creating high-quality, diverse synthetic 3D content often requires expert intervention, making it costly and complex. Recently, efforts to automate this process with learning techniques, particularly diffusion models, have shown significant improvements in tasks like furniture rearrangement. However, applying diffusion models to floor-conditioned indoor scene synthesis remains under-explored. This task is especially challenging as it requires arranging objects in continuous space while selecting from discrete object categories, posing unique difficulties for conventional diffusion methods. To bridge this gap, we present MiDiffusion, a novel mixed discrete-continuous diffusion model designed to synthesize plausible 3D indoor scenes given a floor plan and pre-arranged objects. We represent a scene layout by a 2D floor plan and a set of objects, each defined by category, location, size, and orientation. Our approach uniquely applies structured corruption across mixed discrete semantic and continuous geometric domains, resulting in a better-conditioned problem for denoising. Evaluated on the 3D-FRONT dataset, MiDiffusion outperforms state-of-the-art autoregressive and diffusion models in floor-conditioned 3D scene synthesis. Additionally, it effectively handles partial object constraints via a corruption-and-masking strategy without task-specific training, demonstrating advantages in scene completion and furniture arrangement tasks.

\end{abstract}


\section{Introduction}
Generating diverse and realistic 3D indoor scenes has attracted extensive research in computer vision and graphics. In recent years, this problem has been largely motivated by interior designs of apartments and houses, as well as automatic 3D virtual environment creation for video games and data-driven 3D learning tasks~\cite{Merrell11siggraph-furnitureLayout, Yu11tog-furnitureArrangement, Fisher11siggraph-scenesynthesis, Yu16tvcg-Clutterpalette}. 

To assist these applications, we focus on object-based 3D indoor scene synthesis, which aims at predicting a set of objects with both semantic labels and consistent geometric arrangements. These predicted objects can be rendered into photorealistic scenes using furniture CAD models or other downstream shape prediction methods~\cite{Schult24cvpr-ControlRoom3D}.
State-of-the-art works have recently shown that it is relatively easy to predict reasonable object labels, yet, generating accurate geometric arrangements of objects, \ie predicting realistic positions and orientations, is still very challenging\cite{Paschalidou21neurips-ATISS, Wang21'3DV-SceneFormer}. In particular, as most of the existing approaches are based on autoregressive models, these models tend to quickly start diverging upon a bad prediction in an earlier step.
Therefore, ``de-noising'' approaches that jointly refine  all objects are thus better suited for this type of problems~\cite{Wei23cvpr-LEGONet}.

We propose to adopt diffusion models~\cite{Sohl-Dickstein15icml-diffusion} for this problem, as the iterative denoising scheme is well suited for improving object geometry in 3D scenes. Denoising Diffusion Probabilistic Models (DDPM)~\cite{Ho20neurips-DiffusionModel} adopt Gaussian corruption and denoising of continuous domain data, resulting in efficient training of diffusion models for image synthesis tasks.
Discrete Denoising Diffusion Probabilistic Models (D3PM) \cite{Austin21neurips-D3PM} extended the pioneering work of Hoogeboom \etal~\cite{Hoogeboom21neurips-MultinomialDiffusion} to structured categorical corruption processes for data in discrete state space. Unlike continuous diffusion models, discrete diffusion involves data corruption and recovery using transition probabilities between discrete states. 
Since object attributes are naturally defined on both discrete and continuous domains, we derive a novel mixed discrete-continuous diffusion model, optimally suited for 3D scene synthesis problems. 

A handful of works have recently also adopted diffusion models or iterative denoising strategies for 3D scene synthesis problems~\cite{Inoue23cvpr-LayoutDM, He23arixiv-layoutGeneration, Wei23cvpr-LEGONet, Tang24cvpr-DiffuScene}. 
The closest work to ours is DiffuScene~\cite{Tang24cvpr-DiffuScene}, which generates 3D indoor scenes using DDPM by first converting semantic labels to one-hot encoding vectors and then combining with geometric features in the continuous domain.
While this works overall well, for categorical data, discrete diffusion captures dependencies more naturally and more effectively. Therefore, we propose a novel mixed discrete-continuous denoising formulation which which jointly corruptss and denoises discrete object categories and continuous geometric arrangements, leading to superior results.
Note that we focus on floor-conditioned problems, as these are more practical setups in automatic design of real or virtual worlds. 
We present three technical contributions:
\begin{itemize}
    \item We propose MiDifussion, a novel mixed discrete-continuous diffusion model combining DDPM and D3PM to iteratively denoise objects' semantic and geometric attributes which are naturally defined in discrete and continuous domains, respectively.
    \item We propose a modified version of a time variant transformer denoising network that is compatible with our mixed diffusion formulation and floor plan conditioning.
    \item For applications with partial object constraints, such as scene completion given existing objects, we propose a corruption-and-masking strategy to handle known object conditions without the need to re-train the models.
\end{itemize}
We compare \algorithm against state-of-the-art autoregressive and diffusion models on the common 3D-FRONT~\cite{Fu21iccv-3d-front} dataset, demonstrating that our approach generates more realistic scene layouts, outperforming all existing methods. Code and data are available at \url{https://github.com/MIT-SPARK/MiDiffusion}.

\section{Related Works}
In this section, we review key works in 3D scene synthesis, followed by an overview of diffusion models, and conclude with relevant studies on layout synthesis using diffusion.

\myParagraph{3D Scene Synthesis}
3D scene synthesis is typically achieved by generating a set of object layouts from scratch. 
Classical approaches in the computer graphics community typically employ procedural modeling to apply a set of functions that capture object relationships in indoor~\cite{Yu11tog-furnitureArrangement, Merrell11siggraph-furnitureLayout, Yeh12tog-layoutSynthesis, Qi18cvpr-SceneSynthesis} and outdoor~\cite{Talton11tog-proceduralModeling, Merrell11siggraph-furnitureLayout, Yeh12tog-layoutSynthesis, Prakash19icra-contextAwareSyn, Kar19iccv-MetaSim, Devaranjan20eccv-MetaSim2} settings. 
A handful of methods are based on graph representations of the scene. Meta-Sim~\cite{Kar19iccv-MetaSim} and Meta-Sim2~\cite{Devaranjan20eccv-MetaSim2} learn to modify attributes of scene graphs obtained from probabilistic scene grammars to match synthesized and target images. SG-VAE~\cite{Purkait20eccv-SG-VAE} learns a grammar-based auto-encoder to capture relationships and decode the latent space to a parse tree. 
Most recently, it is common to learn inter-object relationships implicitly without specifying hand-crafted rules. Existing models include feed-forward networks~\cite{Zhang20tog-SceneSynthesis}, VAEs~\cite{Purkait20eccv-SG-VAE, Yang21iccv-SceneSynthesis}, GANs~\cite{Yang21iccv-IndoorSceneGen}, and autoregressive models~\cite{Qi18cvpr-SceneSynthesis, Ritchie19cvpr-SceneSynthesis, Wang19tog-PlanIT, Li19tog-GRAINS, Zhang20tog-SceneSynthesis, Zhou19iccv-SceneGraphNet, Para21iccv-GenerativeLayout, Wang21'3DV-SceneFormer, Paschalidou21neurips-ATISS}. The state-of-the-art autoregressive models~\cite{Wang21'3DV-SceneFormer, Paschalidou21neurips-ATISS} are commonly built with a transformer~\cite{Vaswani17neurips-transformer} backbone, making them capable of modeling object interactions through attention.

\myParagraph{Diffusion Models}
Diffusion probabilistic models \cite{Song20neurips-scoreBasedGenerativeModel, Ho20neurips-DiffusionModel, Austin21neurips-D3PM, Nichol23icml-improvedDiffusionModel, Song21iclr-DDIM} are generative models defined by two Markov processes. The forward process slowly injects random noise to the data, whereas the reverse denoising process recovers the data. 
Conceptually, this approach has connections with denoising score matching methods~\cite{Song19neurips-scoreBasedGenerativeModel, Song21iclr-scoreBasedGenerativeModel}.
Following the pioneering work by \cite{Sohl-Dickstein15icml-diffusion}, diffusion models first gained popularity in 2D image synthesis \cite{Karras22neurips-DiffusionGenerativeModel, Song21iclr-DDIM}, outperforming existing works \cite{Karras20neurips-ImageSynthesisGAN, Esser21cvpr-ImageSynthesis}. 
Starting from Denoising Diffusion Probabilistic Models (DDPM)~\cite{Ho20neurips-DiffusionModel, Sohl-Dickstein15icml-diffusion}, diffusion models typically work with latent variables in the continuous domain. However, it is more natural to represent discrete variables, such as text, in discrete state space. Discrete diffusion models were first applied to text generation in argmax flow~\cite{Hoogeboom21neurips-MultinomialDiffusion}. In the following, Diffusion Probabilistic Models (D3PMs)~\cite{Austin21neurips-D3PM} and VQ-Diffusion~\cite{Gu22cvpr-VQ-Diffusion} have further demonstrated strong results also on character level image generation and text-to-image synthesis.

\myParagraph{Diffusion Models for Layout Synthesis}
Very recently, diffusion models became popular for layout synthesis, including document layout generation~\cite{Inoue23cvpr-LayoutDM, He23arixiv-layoutGeneration}, 3D scene synthesis~\cite{Tang24cvpr-DiffuScene}, furniture re-arrangement~\cite{Wei23cvpr-LEGONet, Tang24cvpr-DiffuScene}, graph-conditioned 3D layout generation~\cite{Zhai23arxiv-CommonScenes, Zhai24eccv-EchoScene}, text conditioned scene synthesis~\cite{Lin24iclr-insturctScene, Fridman23neurips-SceneScape, Gao24cvpr-GraphDreamer}, and 3D scene mesh generation~\cite{Wu24tog-BlockFusion, Meng24arxiv-LT3SD}.
LayoutDM~\cite{Inoue23cvpr-LayoutDM} applies diffusion models in discrete state space. After discretizing position and size to a fixed number of bins, it uses the discrete corruption process by VQ-Diffusion~\cite{Gu22cvpr-VQ-Diffusion} to train the reverse transformer network to predict category, position, and size for document layout generation. Most related to ours, DiffuScene~\cite{Tang24cvpr-DiffuScene} predicts 3D scene layouts using DDPM after converting semantic labels to one-hot encodings, allowing joint diffusion with geometric attributes in the continuous domain. In contrast, our model operates on mixed discrete-continuous domains, leading to more realistic scene layouts and increased robustness. Note that DiffuScene includes a simple demonstration of text-conditioned scene synthesis with a pretrained text parser. However, since we focus on joint semantic and geometric prediction, this line of work~\cite{Feng23neurips-LayoutGPT, Lin24iclr-insturctScene, Schult24cvpr-ControlRoom3D, Hollein23iccv-Text2Room} is not directly comparable to ours.


\section{Preliminary: Diffusion Models}

Diffusion models~\cite{Sohl-Dickstein15icml-diffusion} typically consist of two Markov processes: a forward corruption process and a reverse denoising process. 
The forward process $q(\ContinuousVar_{1:T} | \ContinuousVar_0) = \prod_{t=1}^T q(\ContinuousVar_t | \ContinuousVar_{t-1})$ typically corrupts the data $\ContinuousVar_0 \sim q(\ContinuousVar_0)$ into a sequence of latent variables $\ContinuousVar_{1:T}$ by iteratively injecting controlled noise.
The reverse process $p_{\theta}(\ContinuousVar_{0:T}) = p(\ContinuousVar_T)\prod_{t=1}^Tp_\theta(\ContinuousVar_{t-1} | \ContinuousVar_t)$ progressively denoises the latent variables via a learned denoising network $\theta$. The standard approach to train the denoising network is to minimize the variational bound on the negative log-likelihood:
\begin{equation}
\label{eq:variational_bound}
    \E_{q(\ContinuousVar_0)} {\left[ -\log p_\theta(\ContinuousVar_0) \right]} \leq 
    \E_{q(\ContinuousVar_{0:T})} \left[ -\log \frac{p_\theta(\ContinuousVar_{0:T})}{q(\ContinuousVar_{1:T} | \ContinuousVar_0)}\right] =: L_{vb}
\end{equation}
which can be re-arranged to
\begin{equation}
\label{eq:diffusion_loss}
\begin{split}
    L_{vb} = &\mathbb{E}_{q(\ContinuousVar_0)}[
    \underbrace{D_{\mathrm{KL}}\left(q(\ContinuousVar_T | \ContinuousVar_0)|| p(\ContinuousVar_T)\right)}_{L_T} \\
    & +\sum_{t=2}^T \underbrace{\E_{q(\ContinuousVar_t|\ContinuousVar_0)}[D_{\mathrm{KL}}(q(\ContinuousVar_{t-1} | \ContinuousVar_t, \ContinuousVar_0)|| p_\theta(\ContinuousVar_{t-1} | \ContinuousVar_t))]}_{L_{t-1}} \\
    & + \underbrace{\E_{q(\ContinuousVar_1|\ContinuousVar_0)}[-\log p_\theta(\ContinuousVar_0 | \ContinuousVar_1)]}_{L_0}
    ].
\end{split}
\end{equation}
If the forward process injects known noise, the approximate posterior $q$ has no learnable parameters and hence $L_T$ can be ignored. 
Then, efficient learning of network $\theta$ requires: (1) efficient sampling of $\ContinuousVar_t$ from $q(\ContinuousVar_t|\ContinuousVar_0)$; and (2) tractable computation of $q(\ContinuousVar_{t-1} | \ContinuousVar_t, \ContinuousVar_0)$.

Diffusion models also apply to conditional synthesis problems, such as text conditioned image generation~\cite{Ho20neurips-DiffusionModel, Austin21neurips-D3PM}. Given a conditional input $\CondVar$, we can write the reverse process as $p_{\theta}(\ContinuousVar_{0:T} | \CondVar) = p(\ContinuousVar_T)\prod_{t=1}^Tp_\theta(\ContinuousVar_{t-1} | \ContinuousVar_t, \CondVar)$ and substitute $p_\theta(\ContinuousVar_{t-1} | \ContinuousVar_t)$ by $p_\theta(\ContinuousVar_{t-1} | \ContinuousVar_t, \CondVar)$ in Eq.~\eqref{eq:diffusion_loss}. We drop $\CondVar$ in the remainder of this section for simplicity.

\myParagraph{Continuous State Space}
Denoising Diffusion Probabilistic Model (DDPM)~\cite{ Ho20neurips-DiffusionModel} injects Gaussian noise to a continuous state variable $\ContinuousVar_0$ with fixed variance schedule $\beta_1, \dots, \beta_T \in (0, 1)$:
\begin{equation}
    \label{eq:forward_mc}
    q(\ContinuousVar_t | \ContinuousVar_{t-1}) := \calN(\ContinuousVar_t; \sqrt{1-\beta_t}\ContinuousVar_{t-1}, \beta_t \MI).
\end{equation}
Therefore, there is a closed form expression for sampling $\ContinuousVar_t$ given $\ContinuousVar_0$ at any time $t$:
\begin{equation}
    \label{eq:forward_xt}
    q(\ContinuousVar_t | \ContinuousVar_0) = \calN (\ContinuousVar_t; \sqrt{\Bar{\alpha}_t}\ContinuousVar_0, (1-\Bar{\alpha}_t) \MI),
\end{equation}
where $\alpha_t := 1- \beta_t$ and $\Bar{\alpha}_t := \prod_{s=1}^t \alpha_s$. 
The forward process posteriors are also tractable when conditioned on $\ContinuousVar_0$:
\begin{equation}
    \label{eq:forward_xt_1}
    q(\ContinuousVar_{t-1} | \ContinuousVar_t, \ContinuousVar_0) = \calN(\ContinuousVar_{t-1}; \Tilde{\GaussianMean}_t(\ContinuousVar_t, \ContinuousVar_0), \Tilde{\beta}_t\MI), 
\end{equation}
where $\Tilde{\GaussianMean}_t(\ContinuousVar_t, \ContinuousVar_0) := \frac{\sqrt{\Bar{\alpha}_{t-1}} \beta_t}{1 - \Bar{\alpha}_t} \ContinuousVar_0 + \frac{ \sqrt{\alpha_t}(1-\Bar{\alpha}_{t-1})}{1-\Bar{\alpha}_t}\ContinuousVar_t$ and $\Tilde{\beta}_t := \frac{1-\Bar{\alpha}_{t-1}}{1-\Bar{\alpha}_t} \beta_t$. 
To best approximate Eq.~\eqref{eq:forward_xt_1}, the reverse distributions are Gaussian as well:
\begin{equation}
    \label{eq:reverse_mc}
    p_\networkVar(\ContinuousVar_{t-1} | \ContinuousVar_t) = \calN(\ContinuousVar_{t-1} ; \GaussianMean_\networkVar(\ContinuousVar_t, t), \GaussianCovar_\networkVar(\ContinuousVar_t, t)).
\end{equation}
In practice, the covariance $\GaussianCovar_\theta(\ContinuousVar_t, t) = \sigma^2 \MI$ can be learned, or fixed to $\sigma^2=\Tilde{\beta}_t$ or $\sigma^2=\beta_t$.

\myParagraph{Discrete State Space}
For discrete data, such as semantic labels, it is more natural to define corruption in the discrete domain. 
We denote a scalar discrete variable with $K$ categories by $\discreteVar \in \Int{K}$ and use $\oneHotVec{\discreteVar_t} \in \{0, 1 \}^K$ to represent its one-hot encoding. The variational bound over continuous vector $\ContinuousVar$ in Eq.~\eqref{eq:diffusion_loss} also holds for $\discreteVar$.  For multi-dimensional $\DiscreteVar$, training loss is summed over all elements.
Then the forward process at time $t$ is defined by a transition probability matrix $\DiscreteTransMat_t$, with $[\DiscreteTransMat_t]_{mn} = q(\discreteVar_t =m | \discreteVar_{t-1}=n)$. The categorical distribution over $\discreteVar_t$ given $\discreteVar_{t-1}$ is 
\begin{align}
    q(\discreteVar_{t} | \discreteVar_{t-1}) &:=\text{Cat}(\discreteVar_{t}; p=\DiscreteTransMat_t \oneHotVec{\discreteVar_{t-1}}) \\
    &= \oneHotVec{\discreteVar_t}\tran \DiscreteTransMat_t \oneHotVec{\discreteVar_{t-1}}.
\end{align}
With the Markov property, we can derive
\begin{equation}
    q(\discreteVar_{t} | \discreteVar_{0}) = \oneHotVec{\discreteVar_t}\tran \Bar{\DiscreteTransMat}_t \oneHotVec{\discreteVar_{0}}
\end{equation}
\begin{equation}
    q(\discreteVar_{t-1} | \discreteVar_{t}, \discreteVar_0) = \frac{\left( \oneHotVec{\discreteVar_t}\tran \DiscreteTransMat_t \oneHotVec{\discreteVar_{t-1}} \right) \left( \oneHotVec{\discreteVar_{t-1}}\tran \Bar{\DiscreteTransMat}_{t-1} \oneHotVec{\discreteVar_{0}} \right) }{\oneHotVec{\discreteVar_t}\tran \Bar{\DiscreteTransMat}_t \oneHotVec{\discreteVar_{0}} }
\end{equation}
where $\Bar{\DiscreteTransMat}_t = \DiscreteTransMat_t \DiscreteTransMat_{t-1} \cdots \DiscreteTransMat_1$. 
The denoising network $\theta$ is trained to compute the categorical distributions $p_\theta(\discreteVar_{t-1} | \discreteVar_t)$.
We implement the mask-and-replace strategy by VQ-Diffusion~\cite{Gu22cvpr-VQ-Diffusion}, which improves upon the proposed choices of $\DiscreteTransMat_t$ in~\cite{Austin21neurips-D3PM} by introducing an additional special token \texttt{[MASK]}.
%


\begin{figure*}[!ht]
    \centering
    \includegraphics[width=0.85\linewidth, trim={20 330 50 30},clip]{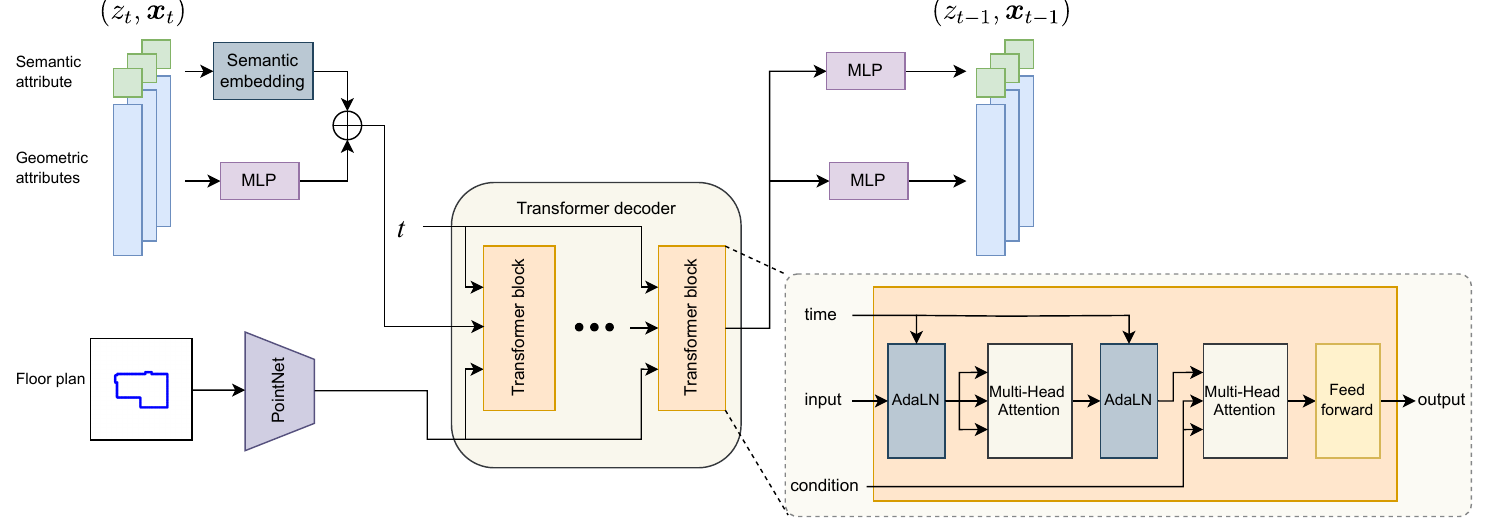}
    \caption{\textbf{Denoising network.} The time-variant decoder backbone takes object features as input, conditioned on the floor plan feature. We utilize the time-variant transformer decoder block from VQ-Diffusion~\cite{Gu22cvpr-VQ-Diffusion}.}
    \label{fig:architecture}
    \vspace{0.1cm}
\end{figure*}

\section{3D Scene Synthesis via Mixed Diffusion}
We present \algorithm, a mixed discrete-continuous diffusion model for 3D indoor scene synthesis. 
We assume each scene $\calS$ is in a world frame with the origin at the center. 
It consists of a floor plan and at most $N$ objects. 
We denote the floor plan by a conditional vector $\CondVar$.
Each object in the scene is characterized by its semantic label $\discreteVar\in \{1, 2, \dots, \numLabels \}$, centroid position $\posVar\in\Real{3}$, bounding box size $\sizeVar\in\Real{3}$, and rotation angle around the vertical axis $\angleVar \in SO(2)$. Inspired by LEGO-Net~\cite{Wei23cvpr-LEGONet}, we represent $\angleVar$ as $[\cos(\gamma), \sin(\gamma)]\tran$ to maintain continuity on $SO(2)$~\cite{Zhou19cvpr-rotationRepresentation}.
We concatenate all geometric attributes to $\ContinuousVar=(\posVar, \sizeVar, \angleVar)$. Then, we can represent a 3D indoor scene as $\calS = (\{( \discreteVar^{i}, \ContinuousVar^{i})\}_{1 \leq i \leq \numObj}, \CondVar$). 
We designate the last semantic label $\discreteVar=\numLabels$ to be an ``empty'' label to work with scenes containing less than $\numObj$ objects and define geometric attributes to all zeros for an ``empty'' object.

The problem of synthesising 3D scenes can be viewed as learning a network $\networkVar$ that maximizes $\log p_\networkVar(\discreteVar_0^i, \ContinuousVar_0^i | \CondVar)$ over all objects and across all scenes, \ie the same variational bound in Eq.~\eqref{eq:diffusion_loss} but over two domains. We drop the superscript $i$ for simplicity.

\subsection{Diffusion in Mixed Discrete-Continuous Domains}
Although object semantic and geometric attributes are in separate domains, we define a corruption process in \algorithm that independently injects domain-specific noises to $\discreteVar_t$ and $\ContinuousVar_t$ as in D3PM and DDPM, respectively:
\begin{equation}
\label{eq:forward_process}
    q \left(\discreteVar_{t}, \ContinuousVar_{t} | \discreteVar_{t-1}, \ContinuousVar_{t-1} \right)  := {\Tilde{q}} (\discreteVar_{t} | \discreteVar_{t-1}) \cdot {\hat{q}} ( \ContinuousVar_{t} | \ContinuousVar_{t-1}).
\end{equation}
This means we can sample $\discreteVar_t$, $\ContinuousVar_t$ independently in training, and factor the posterior distribution as 
\begin{equation}
\label{eq:forward_posterior}
    q \left(\discreteVar_{t-1}, \ContinuousVar_{t-1} | \discreteVar_{t}, \ContinuousVar_{t}, \discreteVar_{0}, \ContinuousVar_{0} \right) =  {\Tilde{q}}(\discreteVar_{t-1} | \discreteVar_{t}, \discreteVar_{0}) \cdot  {\hat{q}} ( \ContinuousVar_{t-1} | \ContinuousVar_{t}, \ContinuousVar_{0}).
\end{equation}
In the backward diffusion process, \Update{we force the network $\theta$ to compute updates by domains conditioned on the previous latent variables:}{we can design a network $\theta$ that computes probability distributions over latent variables by domains to be consistent with the forward process:}
\begin{equation}
\label{eq:backward_process}
\begin{split}
    p_\theta &\left(\discreteVar_{t-1}, \ContinuousVar_{t-1} | \discreteVar_{t}, \ContinuousVar_{t}, \CondVar \right)  \\
    &:=  {\Tilde{p}}_\theta (\discreteVar_{t-1} | \discreteVar_{t}, \ContinuousVar_{t}, \CondVar) \cdot  {\hat{p}}_\theta ( \ContinuousVar_{t-1} | \ContinuousVar_{t}, \discreteVar_{t}, \CondVar).
\end{split}
\end{equation}

Note that the general loss function to train diffusion models in Eq.~\eqref{eq:diffusion_loss} consists of two types: $\{L_{t-1}\}_{2 \leq t \leq T}$, the KL-divergence terms, and $L_0$, a negative log probability. 
The factorization in Eq.~\eqref{eq:forward_process} to \eqref{eq:backward_process} allows us to re-arrange these terms to a pair of domain specific losses:
\begin{equation}
\label{eq:kl_mixed}
\begin{split}
    &L_{t-1}^{mixed} = \E_{q \left(\discreteVar_{t}, \ContinuousVar_{t} | \discreteVar_{0}, \ContinuousVar_{0} \right)}[D_{\mathrm{KL}}(q \left(\discreteVar_{t-1}, \ContinuousVar_{t-1} | \discreteVar_{t}, \ContinuousVar_{t}, \discreteVar_{0}, \ContinuousVar_{0} \right) \\
    & \hspace{4.5cm} || p_\theta \left(\discreteVar_{t-1}, \ContinuousVar_{t-1} | \discreteVar_{t}, \ContinuousVar_{t}, \CondVar \right))] \\
    &= \underbrace{\E_{\Tilde{q}(\discreteVar_t|\discreteVar_0)}[D_{\mathrm{KL}}(\Tilde{q} (\discreteVar_{t-1} | \discreteVar_{t}, \discreteVar_{0}) || \Tilde{p}_\theta (\discreteVar_{t-1} | \discreteVar_{t}, \ContinuousVar_{t}, \CondVar) )]}_{L_{t-1}^{D3PM}} \\
    & \quad + \underbrace{\E_{\hat{q}(\ContinuousVar_t|\ContinuousVar_0)}[D_{\mathrm{KL}}(\hat{q} ( \ContinuousVar_{t-1} | \ContinuousVar_{t}, \ContinuousVar_{0})|| \hat{p}_\theta ( \ContinuousVar_{t-1} | \ContinuousVar_{t}, \discreteVar_{t}, \CondVar))]}_{L_{t-1}^{DDPM}},
\end{split}
\end{equation}
\begin{equation}
\label{eq:l0_mixed}
\begin{split}
    &L_{0}^{mixed} = \E_{q \left(\discreteVar_{1}, \ContinuousVar_{1} | \discreteVar_{0}, \ContinuousVar_{0} \right)} \left[ - \log p_\theta \left(\discreteVar_{0}, \ContinuousVar_{0} | \discreteVar_{1}, \ContinuousVar_{1}, \CondVar \right) \right] \\
    &= \underbrace{\E_{\Tilde{q} \left(\discreteVar_{1} | \discreteVar_{0} \right)} \left[ - \log \Tilde{p}_\theta \left(\discreteVar_{0} | \discreteVar_{1}, \ContinuousVar_{1}, \CondVar \right) \right]}_{L_{0}^{D3PM}} \\
    & \quad + \underbrace{\E_{\hat{q} \left(\ContinuousVar_{1} | \ContinuousVar_{0} \right)} \left[ - \log \hat{p}_\theta \left(\ContinuousVar_{0} | \ContinuousVar_{1}, \discreteVar_{1}, \CondVar \right) \right]}_{L_{0}^{DDPM}} .
\end{split}
\end{equation}
Since $(\ContinuousVar_t, \CondVar)$ is the joint condition term in the reverse step $p_\networkVar$ to predict $\discreteVar_{t-1}$, the first term in Eq.~\eqref{eq:kl_mixed} matches exactly $L_{t-1}$ in D3PM. Similarly, the second term matches $L_{t-1}$ in DDPM. The factorization in Eq.~\eqref{eq:l0_mixed} is similar to that in Eq.~\eqref{eq:kl_mixed}. We provide more details about the factorization step in Eq.~\eqref{eq:kl_mixed} in Appendix~\ref{app:factorization}.

Note that the resulting loss, i.e. Eq.~\eqref{eq:kl_mixed} and \eqref{eq:l0_mixed}, is the exact sum of variational bounds in D3PM and DDPM. The only constraint is that the denoising network has to obey Eq.~\eqref{eq:backward_process}.
Intuitively, summing over discrete and continuous losses is similar to summing losses over each coordinate given multi-dimensional data in vanilla DDPM or D3PM. 
We use a simplified version of $L_{vb}^{DDPM}$ after re-parameterizing the forward process as suggested in DPPM to improve training stability and efficiency.
We add an auxiliary loss, $L_{aux}^{D3PM}$, as proposed in D3PM to encourage good predictions for $\discreteVar_0$ at each time step. 
We include re-parameterization steps as suggested by DDPM and D3PM in Appendix~\ref{app:parameterization}.
The combined loss for \algorithm is:
\begin{equation}
    L^{mixed} = L_{vb}^{DDPM} + L_{vb}^{D3PM} + \lambda L_{aux}^{D3PM}.
\end{equation}

\subsection{Denoising Network}

We use a transformer-based denoising network as shown in Fig.~\ref{fig:architecture}, which takes in latent variables $(\discreteVar_t, \ContinuousVar_t)$ in step $t$ and predicts the categorical distribution $p_\theta(\discreteVar_{t-1} | \discreteVar_{t}, \ContinuousVar_{t}, \CondVar)$, together with the Gaussian mean of $p_\theta ( \ContinuousVar_{t-1} | \ContinuousVar_{t}, \discreteVar_{t}, \CondVar)$ assuming a fixed covariance. At test time, we can simply sample from these distributions to retrieve $(\discreteVar_{t-1}, \ContinuousVar_{t-1})$.

\myParagraph{Feature encoder}
We encode the object features by passing geometric attributes through an MLP and combine them with trainable semantic embeddings.
For each scene, we sample 256 points from a floor plan image along the boundary and compute outward normal vectors. We use the PointNet~\cite{Qi17cvpr-pointnet} floor plan feature extractor from LEGO-Net~\cite{Wei23cvpr-LEGONet} as it is more lightweight and better captures floor boundary than image-based feature extractors. 
\Update{The floor plan feature is then sent to the transformer backbone as a conditioning vector.}{The floor plan feature is then concatenated with the learned position embedding, which is commonly used in transformer architectures, to form the conditioning vector.} 

\myParagraph{Transformer decoder}
The backbone of our denoising network is a time-variant transformer decoder adapted from VQ-Diffusion~\cite{Gu22cvpr-VQ-Diffusion}. Within each transformer block, the time step is injected through the Adaptive Layer Normalization (AdaLN)~\cite{Ba2016arxiv-layerNorm} operator. The conditioning inputs are sent through a multi-head cross-attention layer. 
\Update{Via this transformer decoder, the discrete semantic and continuous geometric features are jointly processed through attention to predict the next diffusion update.}{}


\myParagraph{Feature extractor}
Finally, we feed the output of the transformer decoder to two MLPs to predict a categorical distribution for the semantic labels and Gaussian means for the geometric features respectively.

\subsection{Partial Condition on Objects \label{sec:condition_on_obj}}
3D scene synthesis is sometimes constrained by known object information, as for example in scene completion a given set of furniture pieces. DiffuScene~\cite{Tang24cvpr-DiffuScene} considers these as additional condition inputs to the denoising network, which requires at least a task-specific feature encoder for the conditioning inputs and hence also re-training of the entire model. 
We instead propose to address this issue through a corruption-and-masking strategy which does not require any additional training.
Since the core idea of diffusion models is to recover real data by learning the inverse of a known corruption process, we can force the constrained sections of the latent variables to follow exactly the inverse of the controlled corruption steps in the denoising process.
For example, for scene completion: given $M$ objects, we first compute a sequence of latent variables $\{(\discreteVar_t^i, \ContinuousVar_t^i)\}_{1\leq i\leq M}$ through the corruption process. This allows us to mask out the first $M$ objects using pre-computed latent variables at time $t$ in the reverse process. These latent variables eventually converge to the known input $\{(\discreteVar_0^i, \ContinuousVar_0^i)\}_{1\leq i\leq M}$ without additional training.
This strategy can generalize to other applications, such as furniture arrangement and label conditioned scene synthesis, by targeting different sections of the latent variables. 

\section{Experimental Evaluation \label{sec:experiment}}
In this section, we present evaluation results comparing \algorithm against existing state-of-the-art methods and provide ablation studies, demonstrating the usefulness of our contributions. 
We also show that, without any re-training, our models can be applied to other tasks such as scene completion and furniture arrangement.
Further results and implementation details can be found in the appendix.

\myParagraph{Dataset}
We use the common 3D-FRONT~\cite{Fu21iccv-3d-front} benchmark for evaluation. 3D-FRONT is a synthetic dataset of 18,797 rooms furnished by 7,302 textured objects from 3D-FUTURE~\cite{Fu21ijcv-3d-future}. We evaluate on three room types: 4041 bedrooms, 900 dining rooms, and 813 living rooms. We use the same data processing and train/test data split as~\cite{Paschalidou21neurips-ATISS}. 
We use rotation augmentations in $90^\circ$ increments during training as in ~\cite{Tang24cvpr-DiffuScene} to be consistent with the baselines.

\myParagraph{Baselines}
We compare our proposed approaches against two state-of-the-art 3D scene synthesis methods: (1) ATISS~\cite{Paschalidou21neurips-ATISS}, an autoregressive model with a transformer encoder backbone that generates sequential predictions of objects; (2) DiffuScene~\cite{Tang24cvpr-DiffuScene}, a diffusion model with a conditional 1D U-Net~\cite{Ronneberger15miccai-U-Net} denoising backbone based on DDPM.
For each baseline model we use their open-source code and default hyper parameters, which includes a pretrained ResNet-18~\cite{He16cvpr-ResNet} model as the floor plan feature extractor.
For completeness, we additionally compare \algorithm with DiffuScene in its original application, \ie unconditioned scene synthesis, in Appendix~\ref{app:add_experiment}.

\myParagraph{Implementation}
We train all our models using the Adam optimizer~\cite{Kingma15iclr-adam} with a learning rate of $l_r=2e^{-4}$ and a $0.5$ decay rate every $10k$ for $50k$ epochs on the bedroom dataset, and every $15k$ for $100k$ epochs on the dining room and living room datasets.  We set $\lambda=0.05$ for the auxiliary loss term $L_{aux}^{D3PM}$. More details about implementation and network parameters are presented in Appendix~\ref{app:implementation}.

\myParagraph{Evaluation Metrics}
We randomly select 1000 floor plans from the test set as input and evaluate all models using common metrics from prior works~\cite{Wang21'3DV-SceneFormer, Paschalidou21neurips-ATISS, Tang24cvpr-DiffuScene}: KL divergence of object label distributions (KL$\times 0.01$), Fr\'{e}chet inception distance (FID), Kernel inception distance (KID$\times 0.001$), and classification accuracy (CA\%) following the implementation by ATISS. 
While the KL divergence evaluates the predicted object label distributions, the other metrics are used to compare the layout images with the ground-truth test data.
We render each scene to a 256 x 256 image using CAD models from 3D-FUTURE through top-down orthographic projection. 
Each image captures a 6.2m square for bedrooms, or a 12.2m square living or dining room.
In line with DiffuScene, we color each object according to its semantic class.
Appendix~\ref{app:layout_synthesis} includes example layout images.
We report CA \% over 10 runs with random sampling of predicted layouts for training and testing.
Note that realistic layouts should have a low FID/KID, a low KL and a CA\% that is close to 50\%.
We also compute additional metrics, averaged over all scenes to further evaluate object placement, including the number of generated objects (Obj), percentage of out-of-boundary objects (OOB \%), and bounding box intersection over union (IoU \%).
We dilate the room boundary by 0.1m for OOB \% computation to account for slight under estimation of floor boundary in the raw data.

\subsection{Floor Plan Conditioned 3D Scene Synthesis \label{sec:floor_conditioned}}
In this section, we first compare \algorithm against the state-of-the-art in floor-conditioned 3D scene synthesis experiments without existing objects, before showing several ablations that highlight the impact of our proposed mixed diffusion paradigm.

\begin{figure*}
    \centering
    \begin{minipage}[t]{0.15\textwidth}
       ATISS\\~\cite{Paschalidou21neurips-ATISS}
    \end{minipage}
    \begin{minipage}{.2\textwidth}
        \centering
        \includegraphics[width=0.98\linewidth,trim={80 200 80 220},clip]{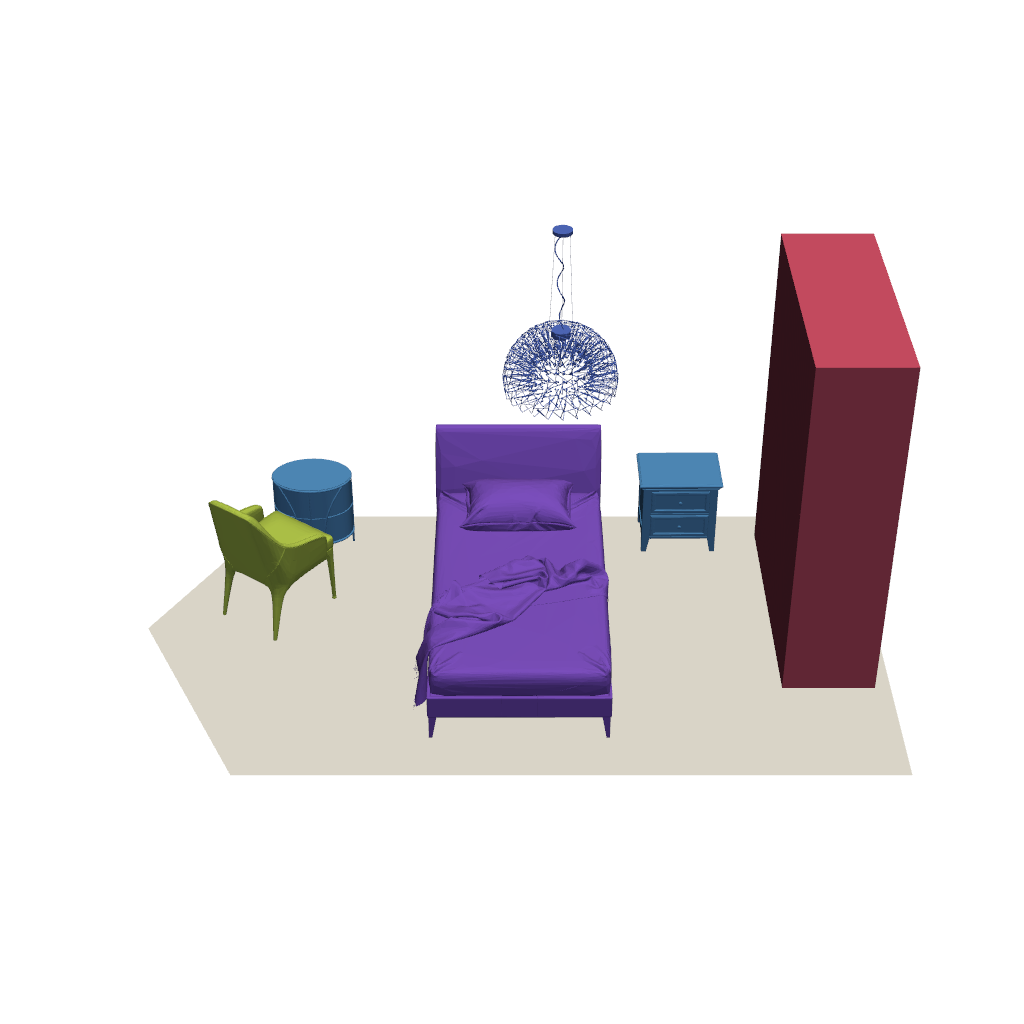}
    \end{minipage}%
    \begin{minipage}{0.2\textwidth}
        \centering
        \includegraphics[width=0.98\linewidth,trim={80 200 80 260},clip]{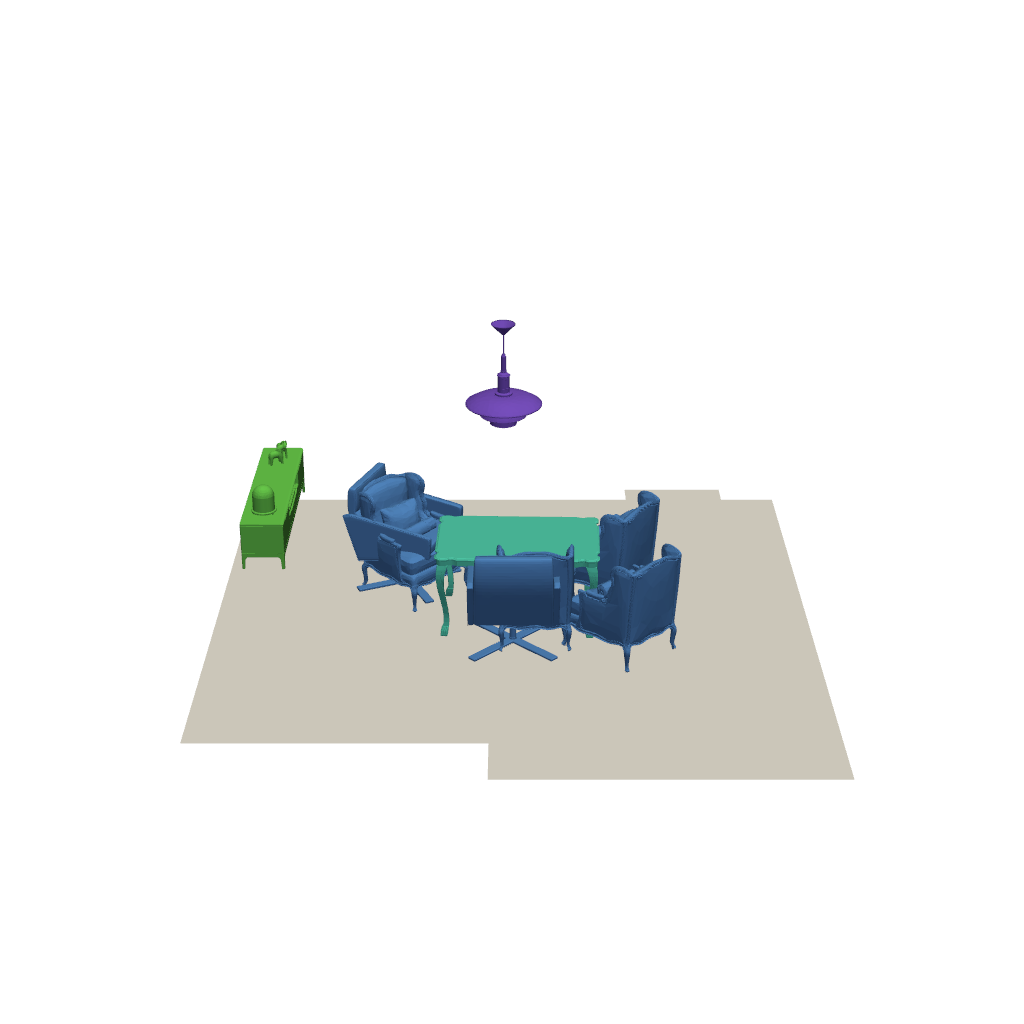}
    \end{minipage}
    \begin{minipage}{.2\textwidth}
        \centering
        \includegraphics[width=0.98\linewidth,trim={40 90 40 220},clip]{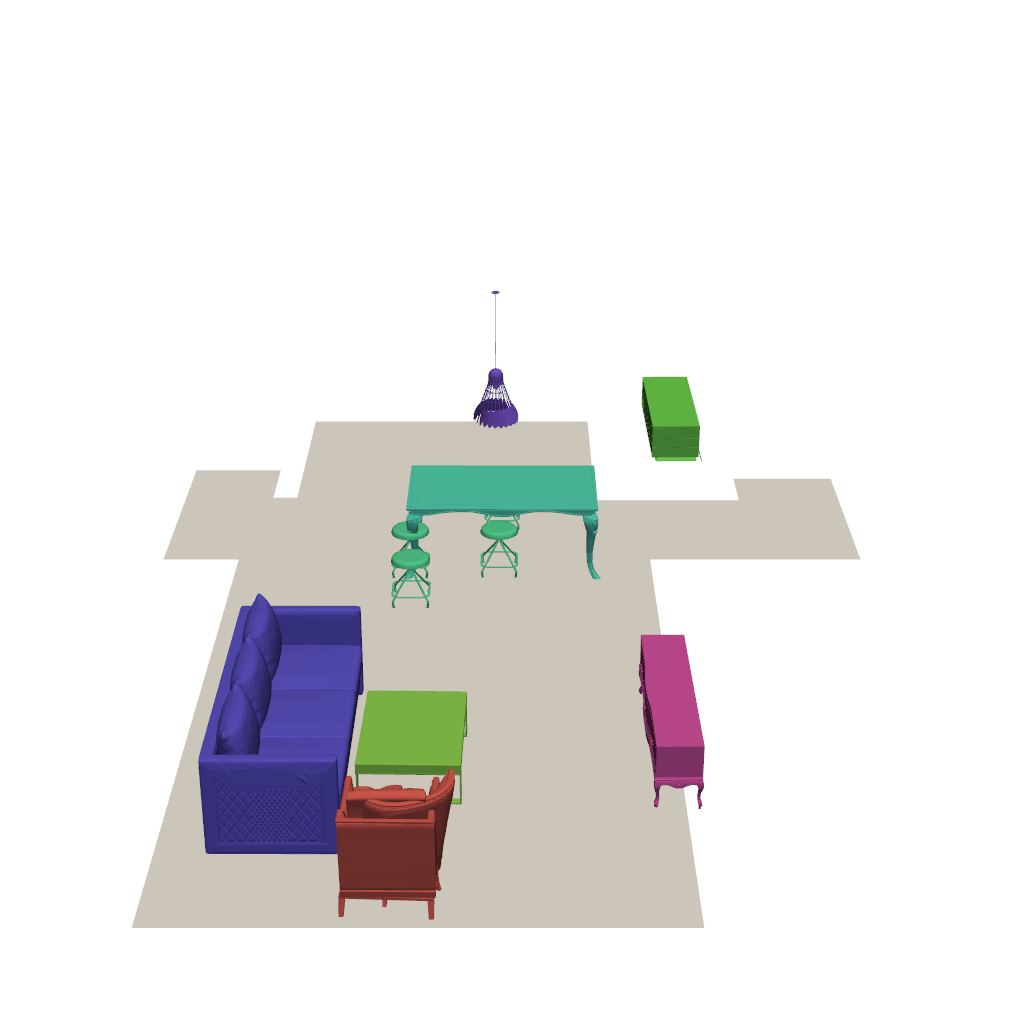}
    \end{minipage}%
    \begin{minipage}{.2\textwidth}
        \centering
        \includegraphics[width=0.98\linewidth,trim={20 120 20 200},clip]{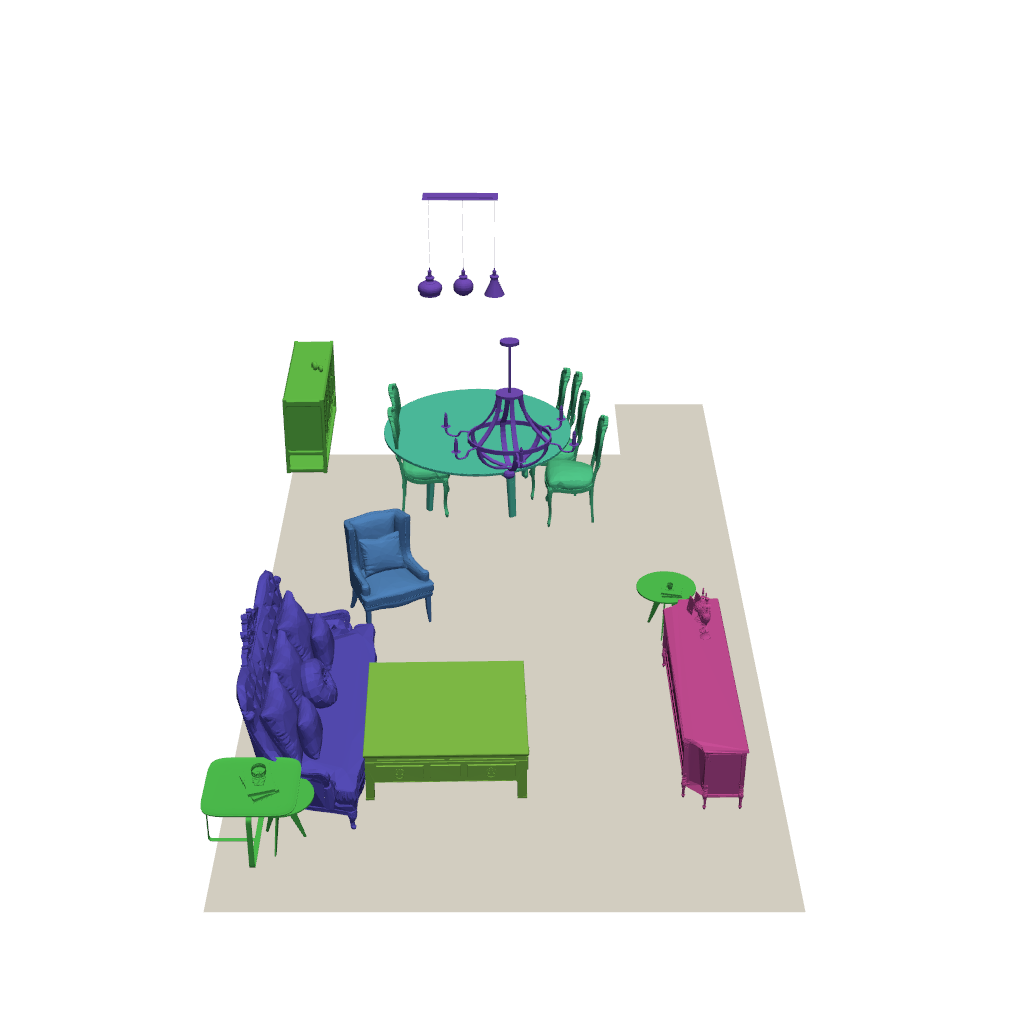}
    \end{minipage}%
    \vspace{-0.1cm}

    \centering
    \begin{minipage}[t]{0.15\textwidth}
       DiffuScene\\~\cite{Tang24cvpr-DiffuScene}
    \end{minipage}
    \begin{minipage}{.2\textwidth}
        \centering
        \includegraphics[width=0.98\linewidth,trim={80 200 80 220},clip]{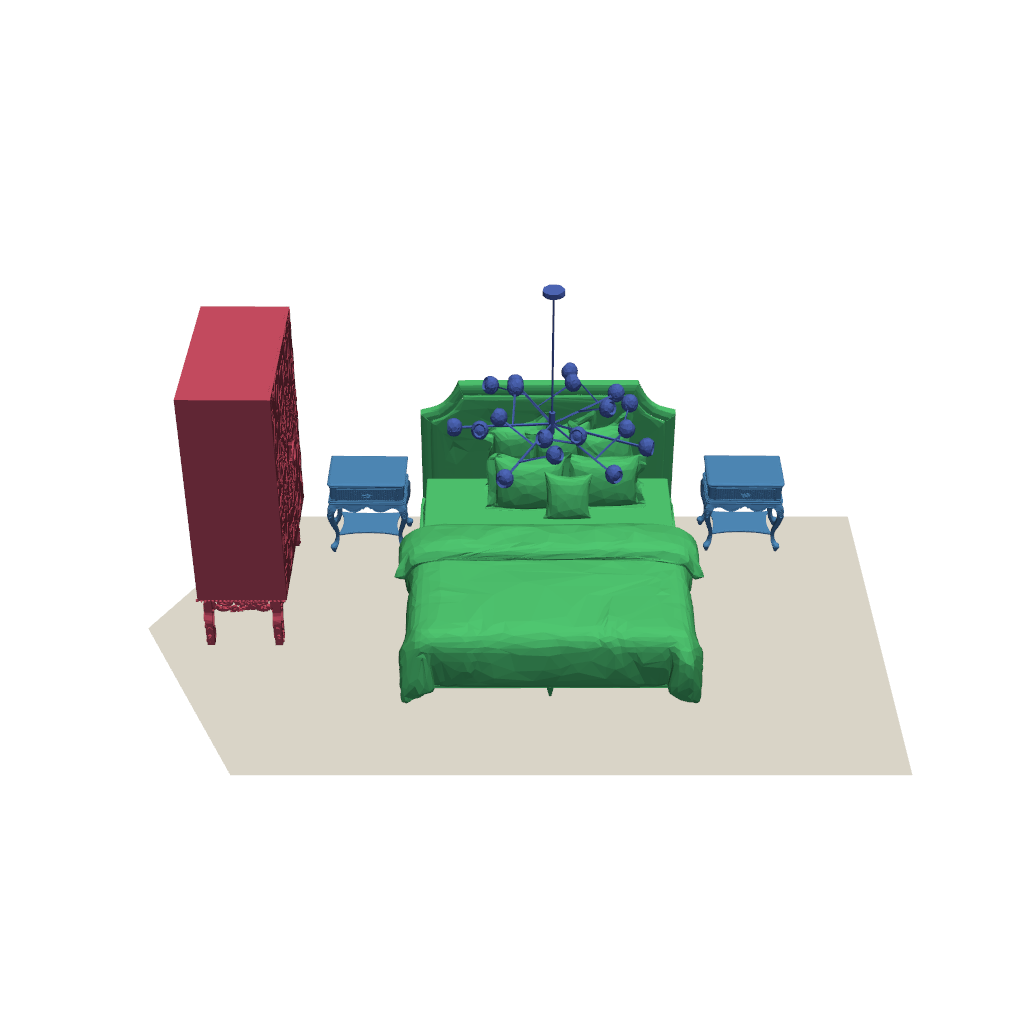}
    \end{minipage}%
    \begin{minipage}{0.2\textwidth}
        \centering
        \includegraphics[width=0.98\linewidth,trim={80 200 80 260},clip]{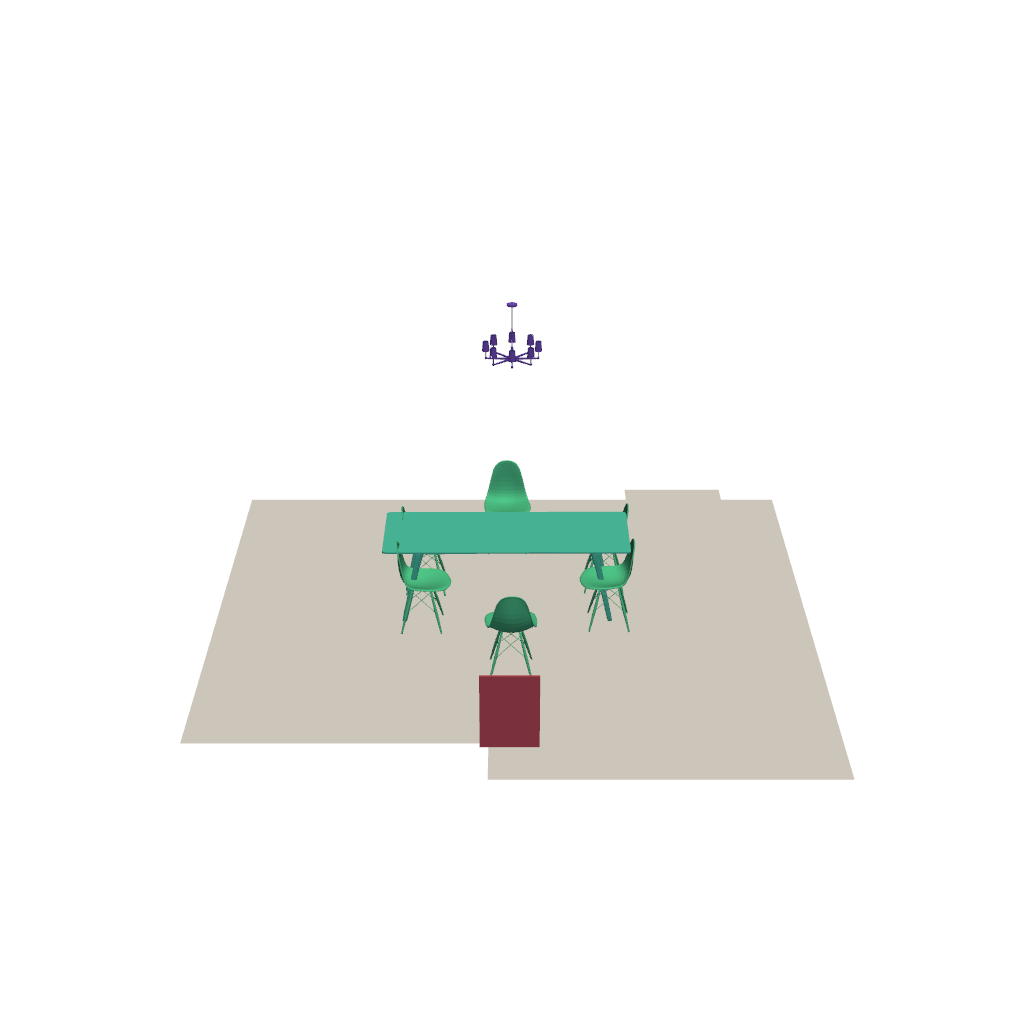}
    \end{minipage}
    \begin{minipage}{.2\textwidth}
        \centering
        \includegraphics[width=0.98\linewidth,trim={40 90 40 220},clip]{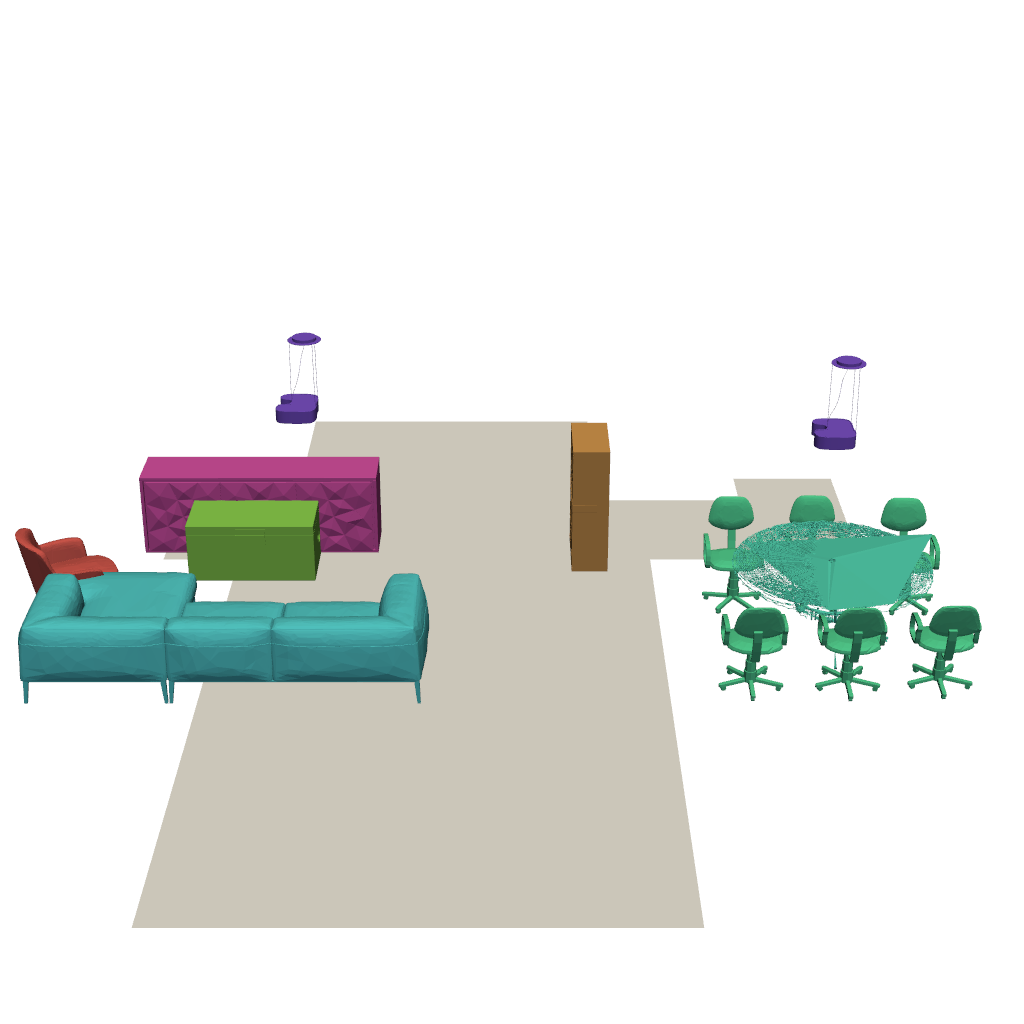}
    \end{minipage}%
    \begin{minipage}{.2\textwidth}
        \centering
        \includegraphics[width=0.98\linewidth,trim={20 120 20 200},clip]{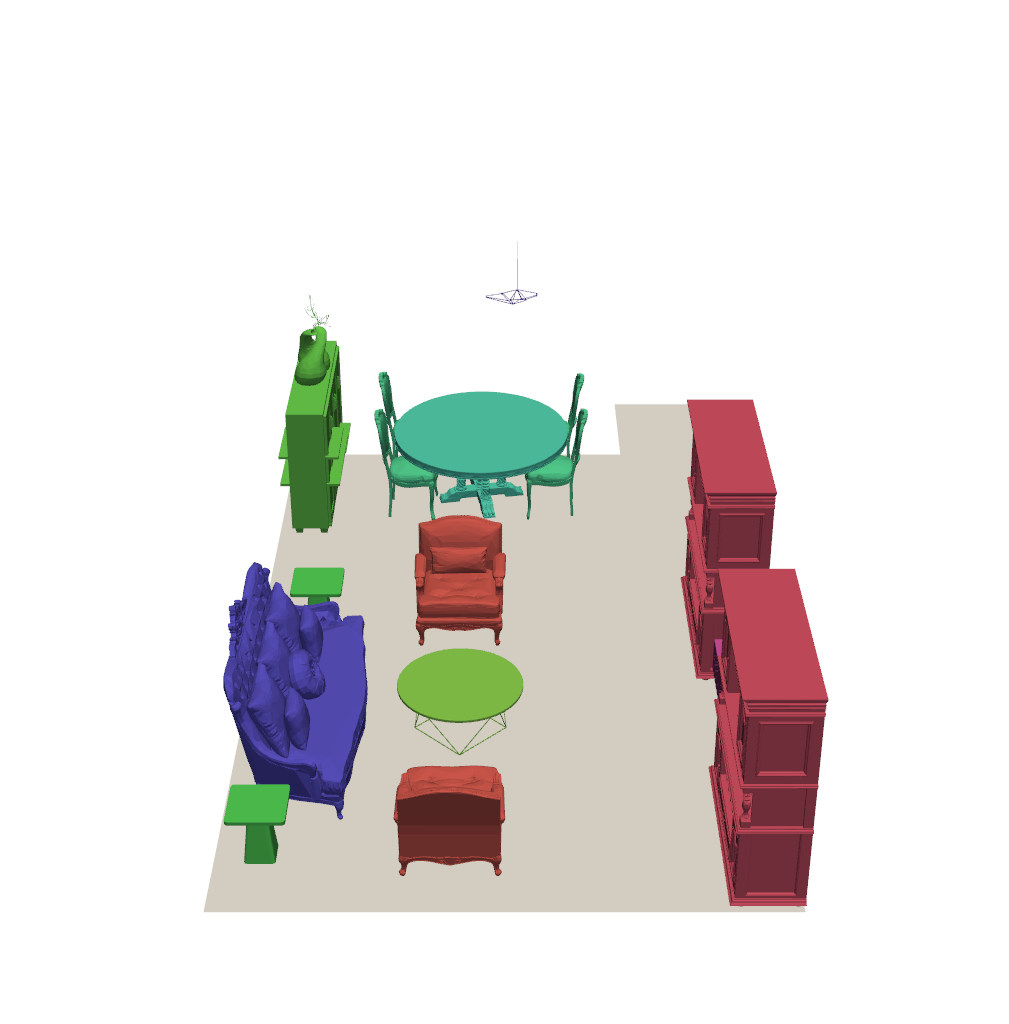}
    \end{minipage}%
    \vspace{0.1cm}
    
    \centering
    \begin{minipage}[t]{0.15\textwidth}
    \algorithm\\
       (Ours)
    \end{minipage}
    \begin{minipage}{.2\textwidth}
        \centering
        \includegraphics[width=0.98\linewidth,trim={80 200 80 220},clip]{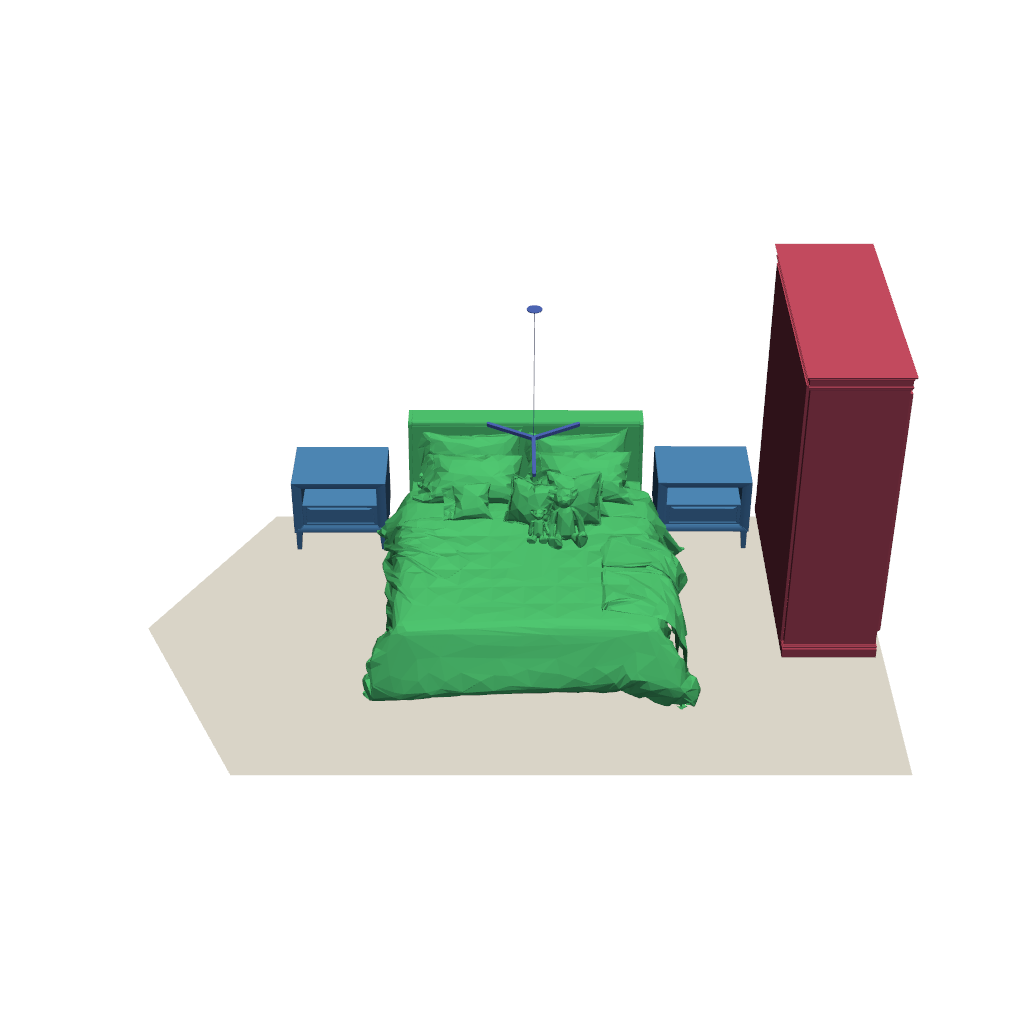}
    \end{minipage}%
    \begin{minipage}{0.2\textwidth}
        \centering
        \includegraphics[width=0.98\linewidth,trim={80 200 80 260},clip]{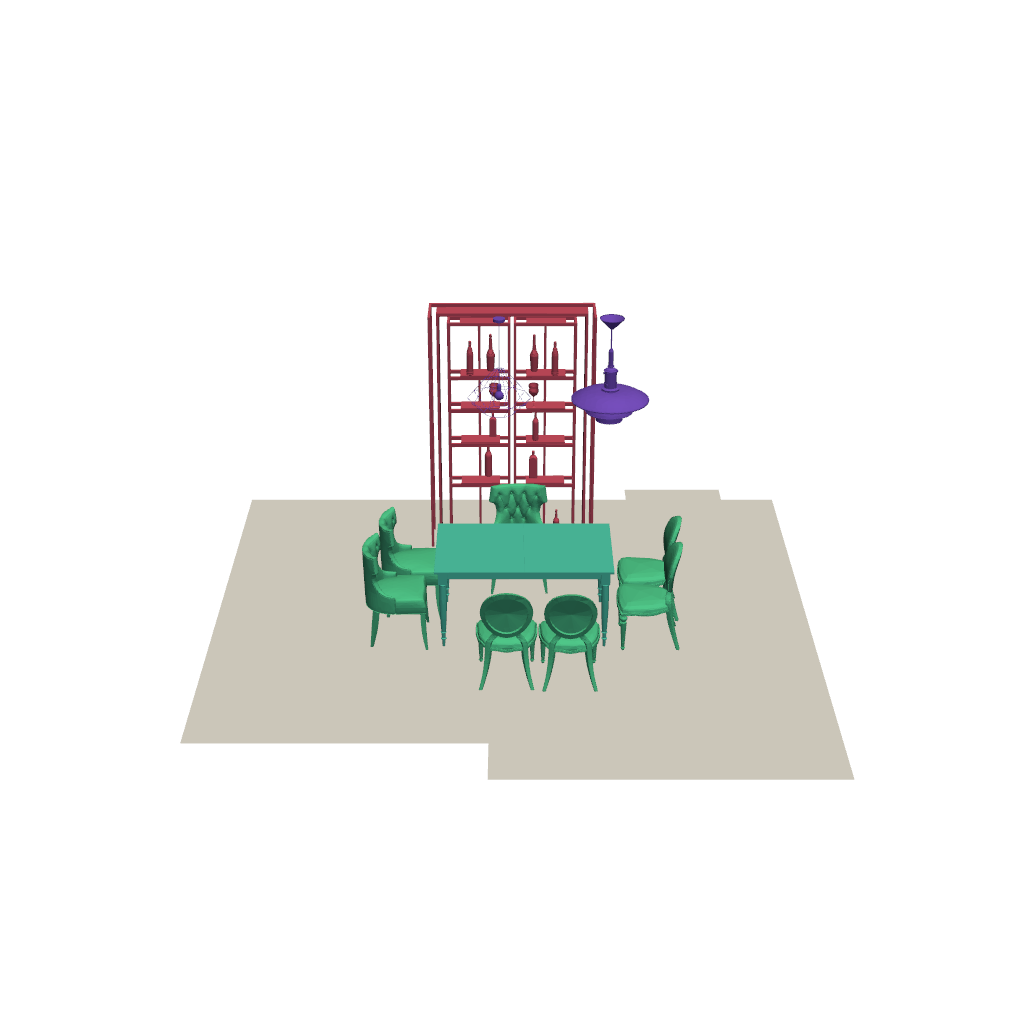}
    \end{minipage}
    \begin{minipage}{.2\textwidth}
        \centering
        \includegraphics[width=0.98\linewidth,trim={40 90 40 220},clip]{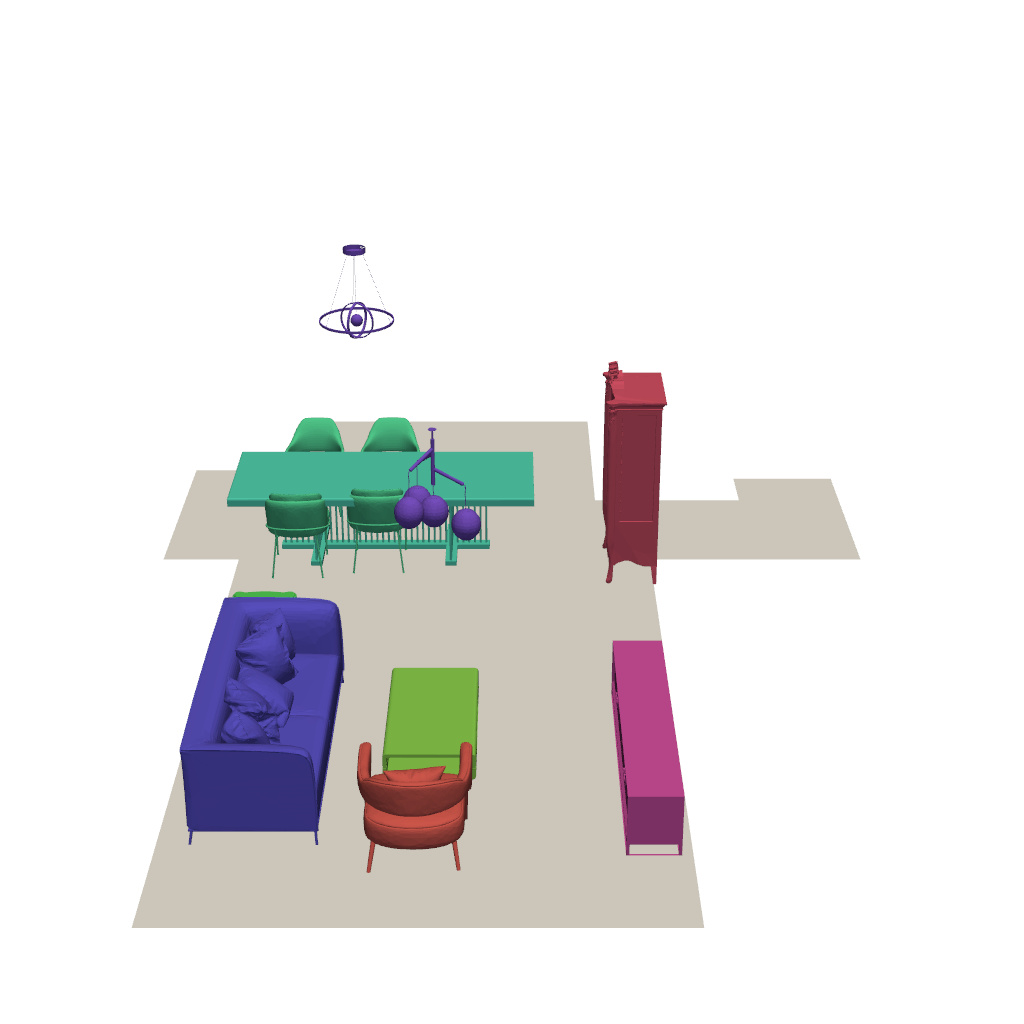}
    \end{minipage}%
    \begin{minipage}{.2\textwidth}
        \centering
        \includegraphics[width=0.98\linewidth,trim={20 120 20 200},clip]{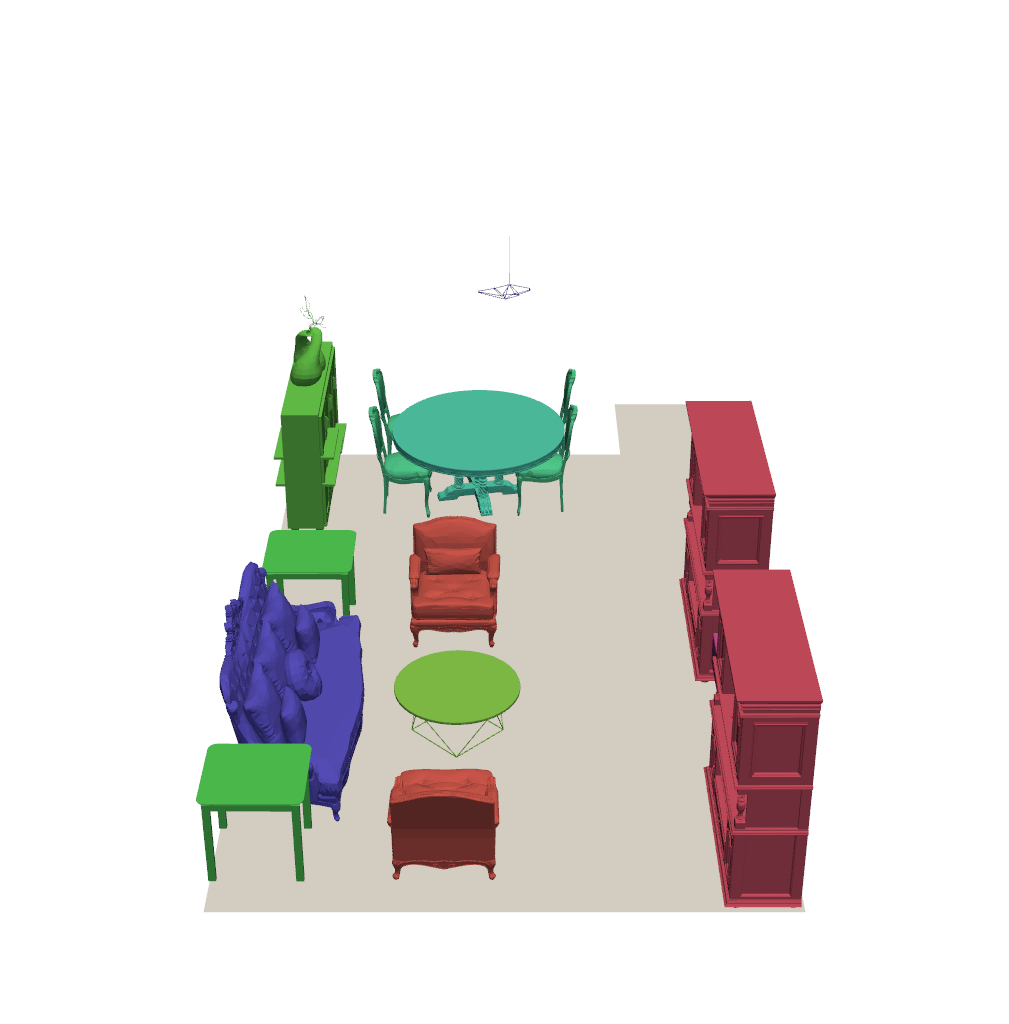}
    \end{minipage}%
    \caption{\textbf{Floor-conditioned scene synthesis}. The meshes are retrieved from the 3D-FUTURE~\cite{Fu21ijcv-3d-future} dataset by size matching within predicted semantic category.  \algorithm generates realistic arrangements while respecting boundary constraints.
    \label{fig:floor_conditioned}}
\end{figure*}

\myParagraph{Comparison with state-of-the-art}
In Fig.~\ref{fig:floor_conditioned}, we present qualitative results rendered with PyVista. 
Note that the object meshes are retrieved from the 3D-FUTURE~\cite{Fu21ijcv-3d-future} dataset by finding the closest match in size given the predicted semantic category. We replace the original textures in the CAD models by a label-specific color for clarity.
We include examples of orthographic projection images which are used for quantitative evaluations in the appendix.
In general, our approach is able to generate realistic object arrangements whilst respecting the boundary constraints. 
ATISS generates slightly less precise object placements compared to the two diffusion based approaches, \eg the table-chairs furniture set in the last example of Fig.~\ref{fig:floor_conditioned} is not exactly symmetric. Both DiffuScene and ATISS show a higher tendency to place objects outside the floor boundary than ours. This problem is more prominent for DiffuScene, probably because the conditional U-Net architecture in the denoising network was not intended for learning boundary constraints.

We show quantitative results in Tables~\ref{tab:conditional} and \ref{tab:conditional_obj}. Overall, our approach achieves substantially better performance in generating realistic layouts, especially in the dining room and living room datasets. There are large margins between our models and the baseline models for FID, KID$\times0.001$ and CA \%, when comparing images of synthesized layouts. 
As for object label distributions, there are no large differences between the individual methods with respect to KL$\times0.01$. 
In Table \ref{tab:conditional_obj}, there is a trade-off between keeping objects inside the boundary (low OOB\%) and maintaining minimal overlap (low IoU\%). 
Although DiffuScene achieves the lowest IoU\%, their models tend to place more objects outside the boundary than other methods, which explains their sub-optimal performance in Table~\ref{tab:conditional}. 
Overall, our approach remains superior in terms of object placement, ensuring consistency between objects and compatibility with floor boundaries.



\begin{table*}[ht]
\resizebox{\textwidth}{!}{
\begin{tabular}{llcccclcccclcccc}
\hline
\multirow{2}{*}{Method} &  & \multicolumn{4}{c}{Bedroom}  &  & \multicolumn{4}{c}{Dining room} &  & \multicolumn{4}{c}{Living room} \\ \cline{3-6} \cline{8-11} \cline{13-16} 
            &  & FID   &KIDx0.001& CA \%   &KLx0.01&  & FID   &KIDx0.001& CA \%   &KLx0.01&  & FID   &KIDx0.001& CA \%   &KLx0.01\\ \hline
ATISS~\cite{Paschalidou21neurips-ATISS}       &  & 65.76         & 1.21          & 54.44 $\pm$ 3.69          & \textbf{0.94} &         & 41.06          & 10.54         & 64.06 $\pm$ 5.63          & 1.89 &         & 38.03          & 9.20          & 63.98 $\pm$ 4.60          & 1.55          \\
DiffuScene~\cite{Tang24cvpr-DiffuScene}  &  & 66.45          & 1.05          & 58.70 $\pm$ 4.93          & 3.41  &        & 45.88          & 9.30          & 65.34 $\pm$ 6.55          & \textbf{1.55}& & 47.91          & 11.75         & 73.96 $\pm$ 4.93          & 1.95          \\
\multicolumn{2}{l}{\algorithm(Ours) }       & \textbf{63.71}   & \textbf{0.24}     & \textbf{53.65 $\pm$ 2.46} & 1.05 & & \textbf{30.63} & \textbf{2.74} & \textbf{52.10 $\pm$ 1.72} & 1.90 &         & \textbf{28.60} & \textbf{1.64} & \textbf{54.04 $\pm$ 3.33} & \textbf{1.54} \\
 \hline
\end{tabular}
}
\vspace{-0.2cm}
\caption{Evaluation results for floor-conditioned 3D scene synthesis. \label{tab:conditional}} 
\vspace{0.1cm}
\end{table*}



\begin{table*}[ht]
\fontsize{7}{8.5}\selectfont
\centering
\begin{tabular}{llccclccclccc}
\hline
\multirow{2}{*}{Method} &  & \multicolumn{3}{c}{Bedroom}  &  & \multicolumn{3}{c}{Dining room} &  & \multicolumn{3}{c}{Living room} \\ \cline{3-5} \cline{7-9} \cline{11-13} 
            &   & Obj           & OOB \%        & IoU \%         &           & Obj            & OOB \%         & IoU \%         & \textbf{} & Obj            & OOB \%        & IoU \%         \\ \hline
ATISS~\cite{Paschalidou21neurips-ATISS}       &   & 5.52          & 16.26         & 1.00         &  & 11.43          & 17.43          & 1.86          &  & 13.00          & 17.39         & 1.49          \\
DiffuScene~\cite{Tang24cvpr-DiffuScene}  &   & 5.05          & 8.06          & \textbf{0.41} &  & 10.90          & 28.16         & \textbf{0.56} &  & \textbf{11.79} & 33.75         & \textbf{0.43} \\
\algorithm(Ours)        &   & \textbf{5.22} & \textbf{3.91} & 0.61          &  & \textbf{10.92} & \textbf{5.77}  & 0.91          &  & 11.91          & \textbf{7.93} & 0.98         \\ \hline
Ground truth          &   & 5.22          & 3.37          & 0.24          &  & 11.11          & 0.73          & 0.48          &  & 11.67          & 1.55          & 0.27  \\ \hline
\end{tabular}
\vspace{-0.2cm}
\caption{Geometric evaluations for floor-conditioned 3D scene synthesis. \label{tab:conditional_obj}} 
\end{table*}

\myParagraph{Ablation study}
We start by ablating our main contribution, i.e. joint discrete-continuous diffusion. To this end, we first compare \algorithm (Mixed + PointNet) to a naive 2-stage approach (2-stage + PointNet).
In the first stage we predict a set of object labels through discrete diffusion, which we then feed to the second stage to estimate their geometric attributes conditioned on the obtained semantic labels. For a fair comparison, we employ the same denoising architecture as in Fig.~\ref{fig:architecture} for both stages and only remove the unused feature encoder or extractor from each stage. The networks are trained independently.
Overall, this 2-stage approach is less efficient and less mathematically principled, and therefore obtains significantly worse results across all metrics compared to \algorithm in Table~\ref{tab:conditional_ablation}. 

Next, we provide three variants of the joint prediction approach using the same transformer decoder backbone to show the effects of mixed diffusion against DDPM, and PointNet floor plan feature extractor against ResNet.
Note that the first variant, DDPM+ResNet, only differs from DiffuScene in its transformer backbone and it already surpasses DiffuScene in Table~\ref{tab:conditional}, suggesting that the transformer architecture is better at learning boundary than its U-Net counterpart.
When using the same floor plan extractor, the mixed diffusion variant always achieves better performance than the DDPM variant, stressing the importance of mixed diffusion. 
Further, while Mixed+ResNet uses the same floor feature extractor as the baselines in Table~\ref{tab:conditional}, it results in better performance.
The more lightweight PointNet floor feature extractor brings a significant performance boost in the dining and living room datasets compared to ResNet, but results in slightly worse performance in the bedroom dataset. 
Since there are much less training scenes for dining and living rooms, and these scenes are larger in physical size, it is likely that PointNet works better under complex and limited training data while suffering from sampling noise during conversion of the floor plan image to boundary points and normals.
Overall, we achieve the best results with our complete formulation (Mixed+PointNet) in which the mixed diffusion approach avoids lifting the discrete inputs to a continuous space as required by the DDPM formulation. We include additional evaluation results in Appendix~\ref{app:add_experiment}



\begin{table*}[!ht]
\resizebox{\textwidth}{!}{
\begin{tabular}{llcccclcccclcccc}
\hline
\multirow{2}{*}{Method} &  & \multicolumn{4}{c}{Bedroom}  &  & \multicolumn{4}{c}{Dining room} &  & \multicolumn{4}{c}{Living room} \\ \cline{3-6} \cline{8-11} \cline{13-16} 
               &  & FID   &KIDx0.001& CA \%   &KLx0.01&  & FID   &KIDx0.001& CA \%   &\small{KLx0.01}&  & FID   &KIDx0.001& CA \%   &KLx0.01\\ \hline

\multicolumn{2}{l}{2-stage + PointNet}  & 65.54	&  1.53	  & 57.83 $\pm$ 4.89   & 3.50      &  & 38.80	& 7.02         & 61.53 $\pm$ 7.85      &    5.08    &  & 35.35	  & 6.30		  & 61.97 $\pm$ 6.55      & 4.73       \\
\hline
\multicolumn{2}{l}{DDPM+ResNet}  & 63.73 &  0.35    & 54.86 $\pm$ 4.22          & 4.72          &  & 31.35          & 3.77          & 53.58 $\pm$ 4.06          & 5.88          &  & 30.44          & 3.05          & 63.48 $\pm$ 6.21          & 6.06          \\
\multicolumn{2}{l}{\Update{Mixed+ResNet}}{}   & \Update{63.88}{}        & \Update{0.20}{}      & \Update{54.41 $\pm$ 4.03}{}          & \Update{2.66}{}          &  & \Update{31.29}{}          & \Update{3.81}{}        & \Update{53.41 $\pm$ 3.48}{} & \Update{2.04}{}  &  & \Update{30.15}{}  & \Update{2.81}{}   & \Update{57.21 $\pm$ 3.97}{}          & \Update{1.74}{}          \\ 
\multicolumn{2}{l}{DDPM+PointNet}  & 65.25          & 1.25          & 57.00 $\pm$ 4.67          & 2.14          &  & 31.43          & 2.85        & 52.71 $\pm$ 4.55          & 2.72          &  & 28.63          & 1.68          & 55.27 $\pm$ 5.44          & 1.93          \\ \hline
\multicolumn{2}{l}{\textbf{Mixed+PointNet}}  & \textbf{63.71}        & \textbf{0.24}     & \textbf{53.65 $\pm$ 2.46} & \textbf{1.05} & & \textbf{30.63} & \textbf{2.74} & \textbf{52.10 $\pm$ 1.72} & \textbf{1.90} &         & \textbf{28.60} & \textbf{1.64} & \textbf{54.04 $\pm$ 3.33}  & \textbf{1.54} \\
 \hline
\end{tabular}
}
\vspace{-0.2cm}
\caption{Ablation study for floor-conditioned 3D scene synthesis. \label{tab:conditional_ablation}} 
\end{table*}

\myParagraph{Diversity}
We study the diversity of the predicted layouts by measuring the average standard deviation (std) of object centroid positions and bounding box sizes. Since vertical positions are highly correlated with semantic labels, which are already evaluated through KL-divergence, we compute the average std over the two planar axes only. The results are shown in Table~\ref{tab:diversity}, where the ``IB'' suffix means results are computed over ``in-boundary'' objects only to remove bad object placement.
All three approaches are quite similar in terms of diversity.
Overall, our approach generates the most diverse object placements that are within the floor boundary (\ie Position-IB), but is slightly worse than DiffuScene in size variety.

\begin{table*}[ht]
\fontsize{7}{8.5}\selectfont
\resizebox{\textwidth}{!}{
\begin{tabular}{llcccclcccclcccc}
\hline
\multirow{2}{*}{Method} &  & \multicolumn{4}{c}{Bedroom}  &  & \multicolumn{4}{c}{Dining room} &  & \multicolumn{4}{c}{Living room} \\ \cline{3-6} \cline{8-11} \cline{13-16} 
            &  & Position       & Position-IB    & Size           & Size-IB                       &  & Position       & Position-IB    & Size           & Size-IB              &  & Position       & Position-IB    & Size           & Size-IB              \\ \hline
ATISS~\cite{Paschalidou21neurips-ATISS}       &  & \textbf{1.074} & 1.035          & 0.709          & 0.663                         &  & 1.586          & 1.504          & 0.434          & 0.385                &  & 1.730          & 1.663          & 0.442          & 0.409                \\
DiffuScene~\cite{Tang24cvpr-DiffuScene}  &  & 1.073         & 1.059          & \textbf{0.718} & \textbf{0.691}                &  & \textbf{1.596} & 1.427          & \textbf{0.454} & \textbf{0.423}       &  & \textbf{1.752} & 1.575          & \textbf{0.482} & \textbf{0.439}       \\
\multicolumn{2}{l}{\algorithm(Ours) } & 1.073          & \textbf{1.067} & 0.698          & 0.680                         &  & 1.568          & \textbf{1.558} & 0.414          &  0.399               &  & 1.668          & \textbf{1.667} & 0.439          & 0.428                 \\ \hline
\multicolumn{2}{l}{Ground truth} & 1.091	       & 1.082	 & 0.722	      & 0.690                         &  & 1.623	    & 1.618	& 0.450	      &  0.447	   &  & 1.735	  &1.729	 & 0.463 & 	0.460                  \\
 \hline
\end{tabular}
}
\vspace{-0.2cm}
\caption{Average std of predicted object positions and sizes for floor-conditioned 3D scene synthesis. \label{tab:diversity}} 
\end{table*}


\begin{table*}[ht]
\resizebox{\textwidth}{!}{
\begin{tabular}{llcccccccccccccc}
\cline{1-16}
                           & \multirow{2}{*}{Method} & \multicolumn{4}{c}{Bedroom}                                     &  & \multicolumn{4}{c}{Dining room}                                &  & \multicolumn{4}{c}{Living room}                                \\ \cline{3-6} \cline{8-11} \cline{13-16} 
                           &                         & FID            & KID x 0.001    & KL x 0.01     & OOB \%        &  & FID            & KID x 0.001   & KL x 0.01     & OOB \%        &  & FID            & KID x 0.001   & KL x 0.01     & OOB \%        \\ 
\cline{1-16} 
\multirow{3}{*}{\rotatebox[origin=c]{90}{1 obj}}  & ATISS~\cite{Paschalidou21neurips-ATISS}                   & 60.59          & 0.19           & 1.04          & 13.52         &  & 36.63          & 7.77          & 1.79          & 15.30         &  & 34.34          & 7.15          & 1.46          & 16.19         \\
                           & DiffuScene-SC~\cite{Tang24cvpr-DiffuScene}           & 83.90          & 19.17          & 18.34         & \textbf{3.11} &  & 34.12          & 4.82          & 1.59          & 12.13         &  & 36.32          & 6.64          & \textbf{1.30} & 21.74         \\
                           & MiDiffusion (ours)      & \textbf{60.08} & \textbf{-0.03}  & \textbf{0.91} & 3.32          &  & \textbf{28.56} & \textbf{1.88} & \textbf{1.58} & \textbf{5.29} &  & \textbf{27.86} & \textbf{1.98} & 1.49          & \textbf{7.81} \\ 
\cline{1-16} 
\multirow{3}{*}{\rotatebox[origin=c]{90}{3 obj}} & ATISS~\cite{Paschalidou21neurips-ATISS}                   & 51.83          & -0.57          & \textbf{0.66} & 8.89          &  & 33.53          & 5.39          & 1.60          & 12.26         &  & 31.78          & 5.39          & 1.43          & 13.03         \\
                           & DiffuScene-SC~\cite{Tang24cvpr-DiffuScene}            & \textbf{42.77} & \textbf{-1.08} & 0.80          & 5.13          &  & 30.58          & 3.67          & \textbf{0.84} & 10.04         &  & 32.07          & 4.58          & \textbf{0.57} & 16.13         \\
                           & MiDiffusion (ours)      & 52.19          & -0.32          & 1.00          & \textbf{4.05} &  & \textbf{27.60} & \textbf{1.85} & 1.72          & \textbf{3.97} &  & \textbf{27.45} & \textbf{1.91} & 0.98          & \textbf{6.76} \\ 
\hline 
\end{tabular}
}
\vspace{-0.2cm}
\caption{Evaluation results for scene completion. \label{tab:scene_completion}} 
\end{table*}



\begin{table*}[!ht]
\centering
\resizebox{0.9\textwidth}{!}{
\begin{tabular}{llcccclcccclcccc}
\hline
\multirow{2}{*}{Method} &  & \multicolumn{4}{c}{Bedroom}  &  & \multicolumn{4}{c}{Dining room} &  & \multicolumn{4}{c}{Living room} \\ \cline{3-6} \cline{8-11} \cline{13-16} 
            &  & FID   &KIDx0.001& OBB \%   &IoU \% &  & FID   &KIDx0.001& OBB \%   &IoU \% &  & FID   &KIDx0.001& OBB \%   &IoU \% \\ \hline
DiffuScene-FA~\cite{Tang24cvpr-DiffuScene}  &  & 61.03          & 0.14          & 8.03          & 0.75  &        & 33.90          & 4.14          & 12.30          & 1.16  & & 34.62          & 6.04         & 17.84          & 1.31          \\
\multicolumn{2}{l}{MiDiffusion (ours)}  & \textbf{59.08}   & \textbf{-0.24}     & \textbf{7.34} & \textbf{0.55} & & \textbf{31.72} & \textbf{3.91} & \textbf{6.35} & \textbf{0.96} &         & \textbf{30.02} & \textbf{3.64} & \textbf{8.33} & \textbf{0.98} \\
 \hline
\end{tabular}
}
\vspace{-0.2cm}
\caption{Evaluation results for furniture arrangement experiment. \label{tab:furniture}} 
\end{table*}

\subsection{Floor Plan Conditioned 3D Scene Synthesis with Object Constraints\label{sec:exp_object_condition}}
Recall from Sec.~\ref{sec:condition_on_obj} that we can re-use the same \algorithm models for problems involving partial object constraints through a corruption-and-masking strategy. This approach forces the relevant object attributes to follow the reverse of a pre-computed corruption process in the denoising step, without the need of task-specific retraining. Although it is also feasible to directly mask out relevant features by known inputs in each denoising step as proposed in~\cite{Lin24iclr-insturctScene}, this approach creates out-of-distribution latent variables in the intermediate denoising step and is therefore less principled. We show comparison between the two masking approaches and additional applications in Appendix.

\subsubsection{Scene completion}


\begin{figure}[ht]
    \centering
    \begin{subfigure}{0.4\linewidth}
        \centering
        \includegraphics[width=\linewidth,trim={60 180 160 230},clip]{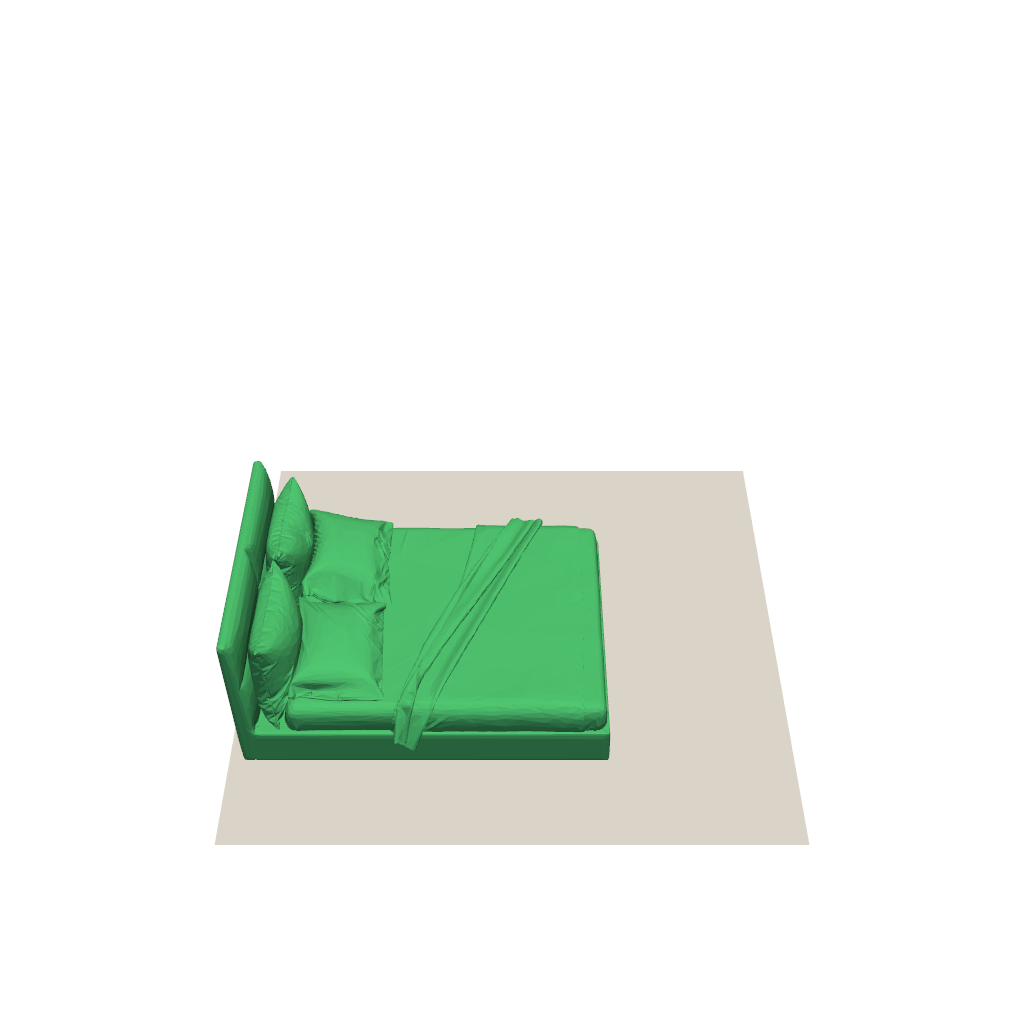}
    \end{subfigure}
    \hfill
    \begin{subfigure}{0.4\linewidth}
        \centering
        \includegraphics[width=\linewidth,trim={160 180 60 230},clip]{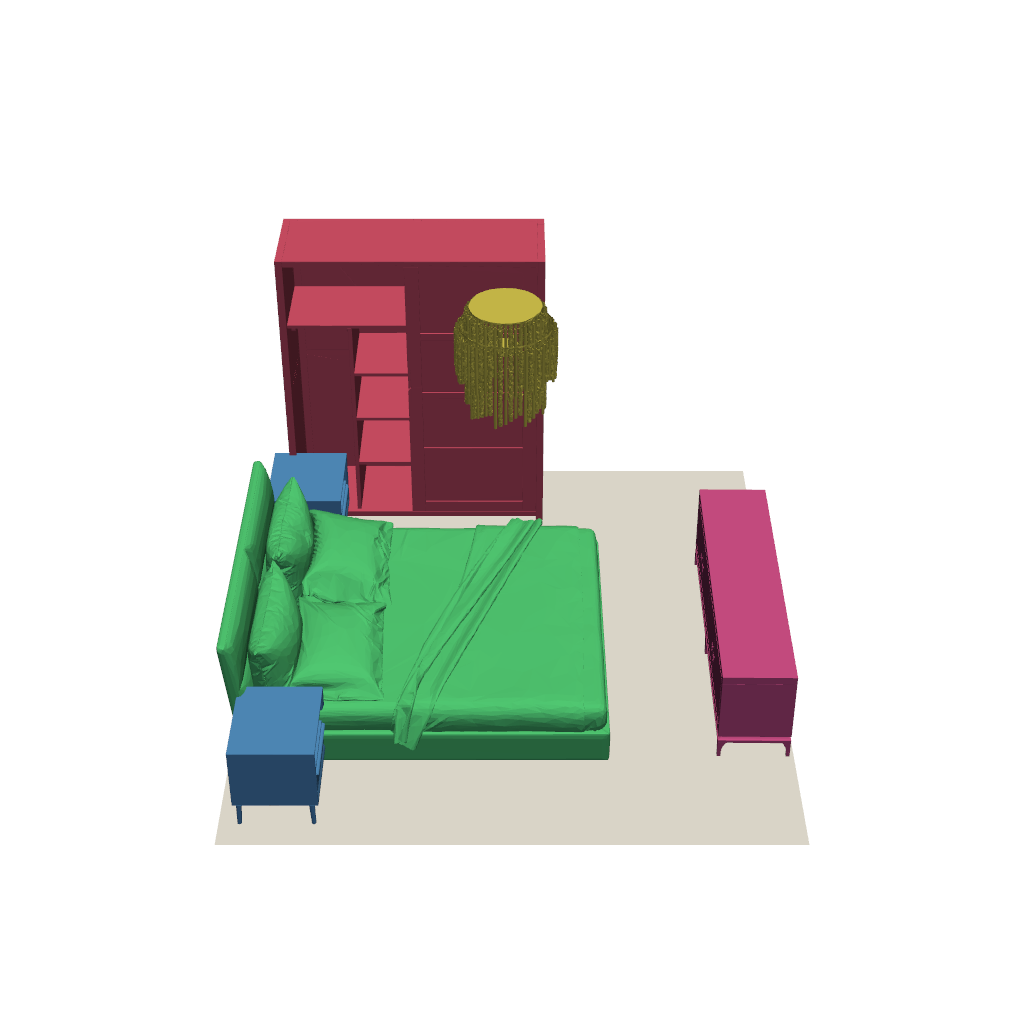}
    \end{subfigure}
    
    \vspace{0.2cm}
    \begin{subfigure}{0.45\linewidth}
        \centering
        \includegraphics[width=\linewidth,trim={60 180 100 240},clip]{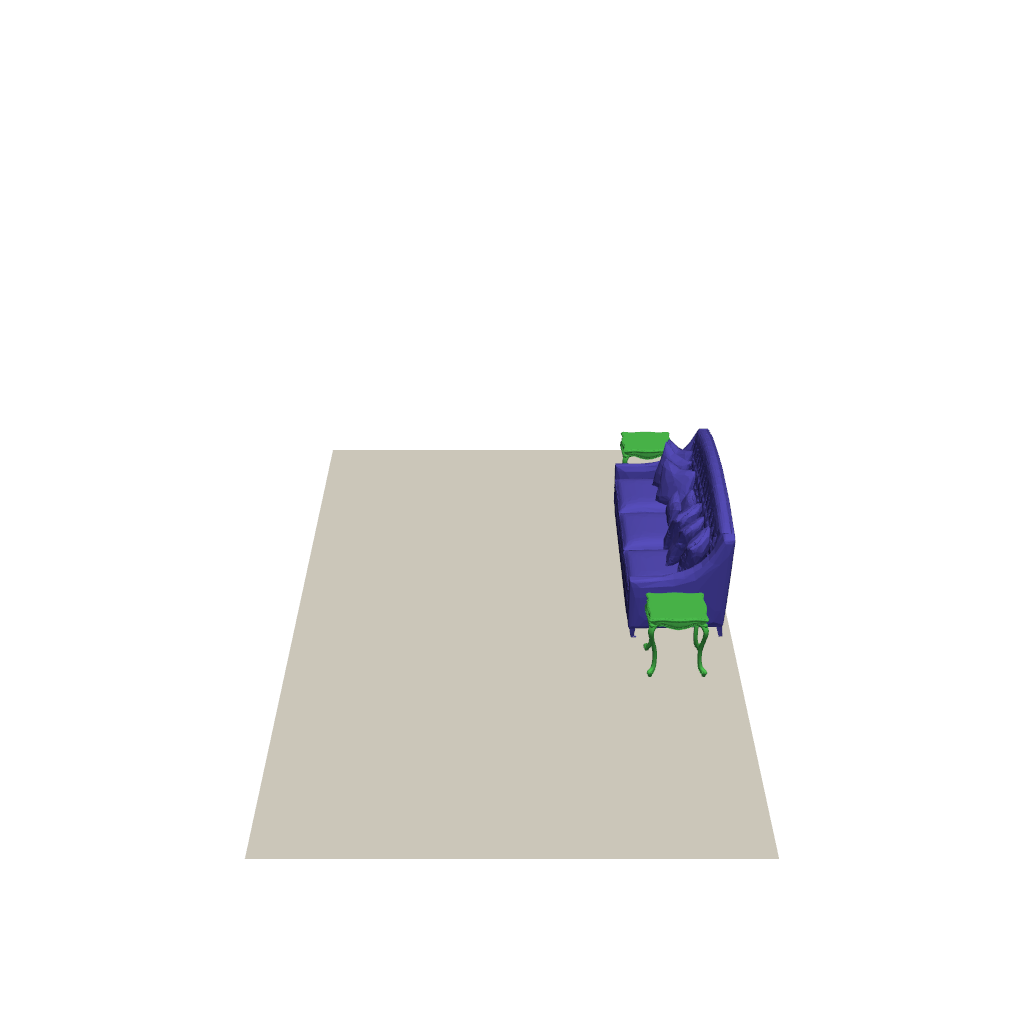}
    \end{subfigure}
    \hfill
    \begin{subfigure}{0.45\linewidth}
        \centering
        \includegraphics[width=\linewidth,trim={100 180 60 240},clip]{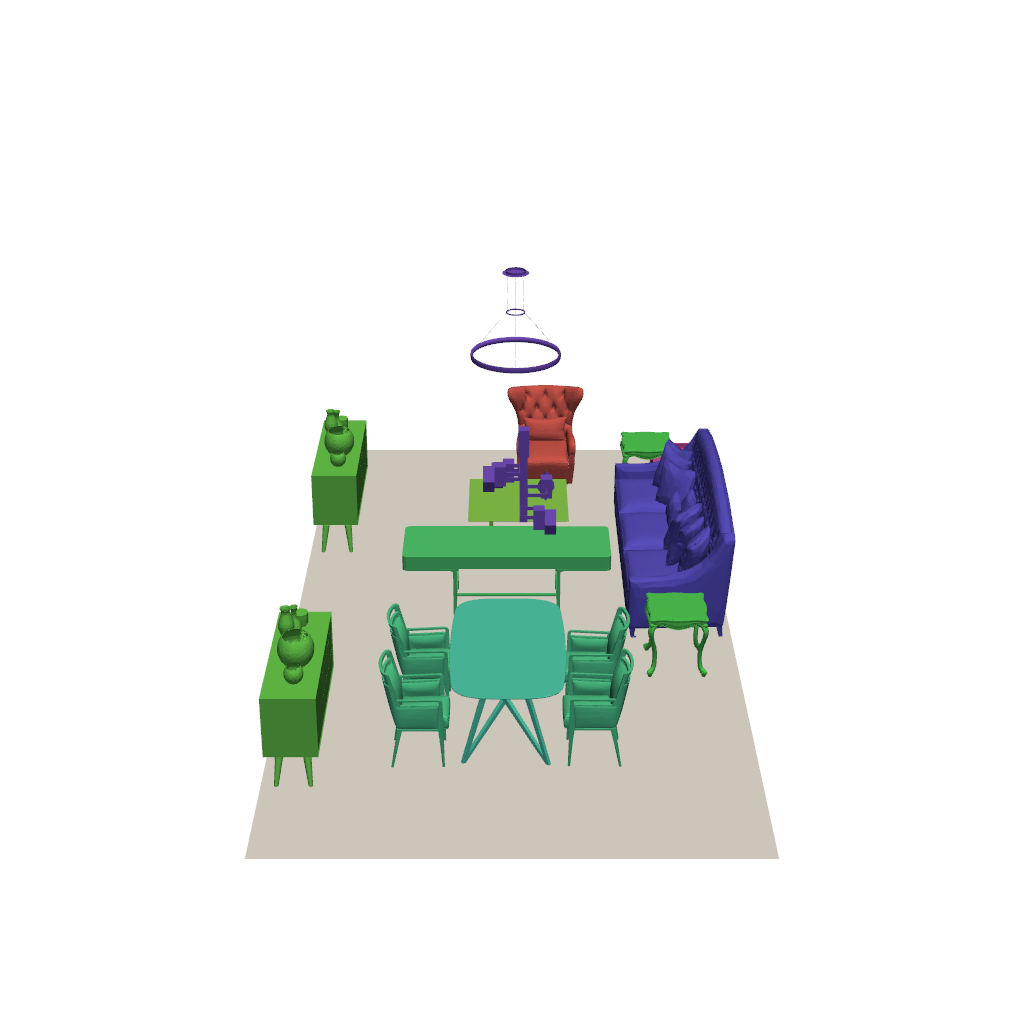}
    \end{subfigure}
    
    \caption{\textbf{Scene completion}. Bedroom (top) and living room (bottom) scene completion examples.}
    \label{fig:completion}
\end{figure}

We first compare our approach with baselines for scene completion (\ie scene synthesis conditioned on existing objects). We train the DiffuScene-SC variant, which includes an additional module to learn existing object features as a conditional input, using their suggested hyper-parameters and providing 3 existing objects during training. The number of existing objects can be arbitrary at test time.
We use the same models for ATISS and \algorithm. 
We use the proposed corruption-and-masking strategy over \algorithm models trained in Section 5.1.
Fig.~\ref{fig:completion} shows a simple example of completing a bedroom design that originally only consists of a single bed and another more complex example for completion of a living space. Note that \algorithm is able to generate a natural and symmetric set of furniture layout.

We show quantitative evaluations in Table \ref{tab:scene_completion}.
This is an easier problem for ATISS and DiffuScene than the previous problem. In particular, ATISS is autoregressive and therefore existing objects allow the models to skip the first few iterations reducing the chance for mistakes. DiffuScene-SC is, on the other hand, specifically designed and trained for this task. Compared to results in Table~\ref{tab:conditional}, these approaches indeed benefit from using existing objects as anchors.
Nevertheless, without any retraining, our approach still outperforms these baselines when given only 1 existing object. When given 3 objects, our models are still better on the dining room and living room datasets, however, are outperformed by DiffuScene-SC on the bedroom dataset. Note that DiffuScene-SC is particularly optimized for the 3 objects scenario and that the bedroom data has only an average of only 5.22 objects. Hence, specifically trained networks, such as DiffuScene-SC, might be better suited given the very high percentage of pre-existing objects.

\subsubsection{Furniture Arrangement}


\begin{figure}[ht]
    \centering
    \vspace{-0.3cm}
    \begin{subfigure}{0.4\linewidth}
        \centering
        \includegraphics[width=\linewidth,trim={20 60 20 130},clip]{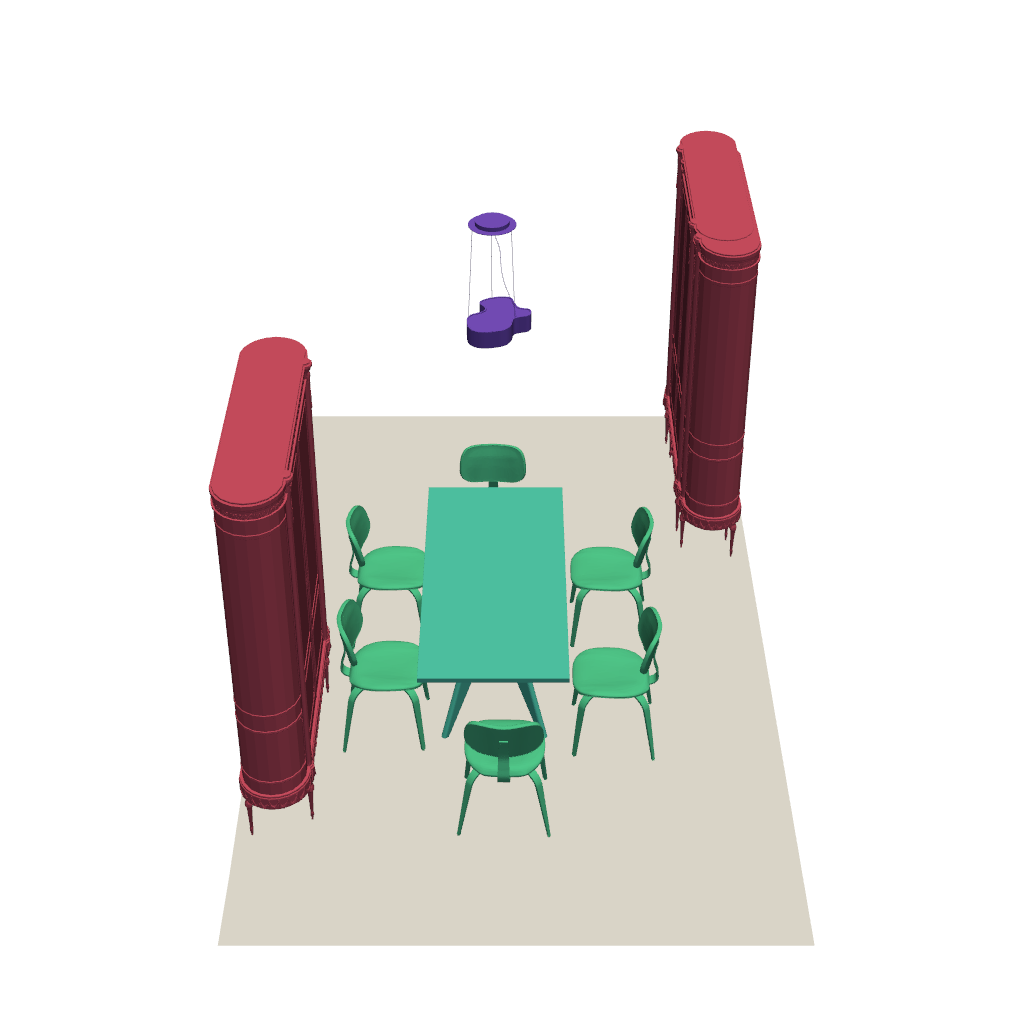}
    \end{subfigure}
    \hfill
    \begin{subfigure}{0.4\linewidth}
        \centering
        \includegraphics[width=\linewidth,trim={20 60 20 130},clip]{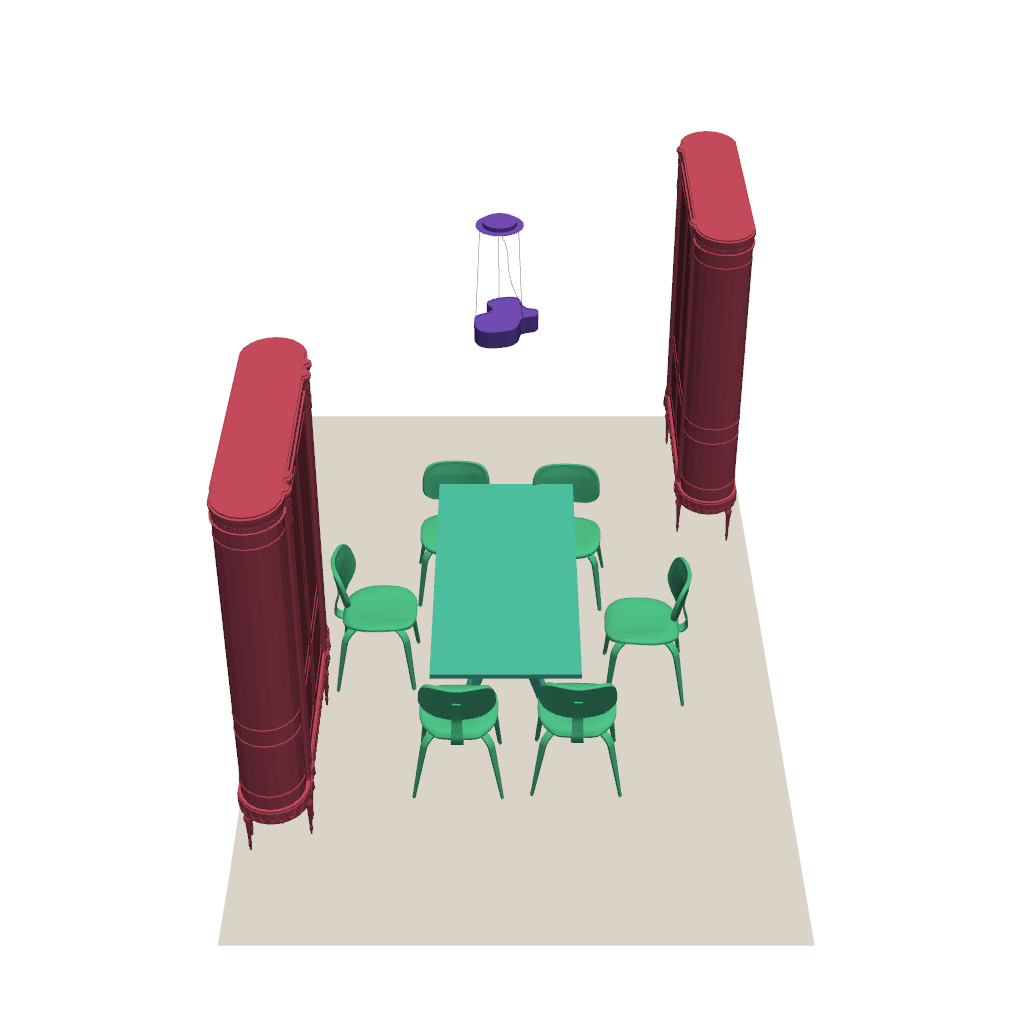}
    \end{subfigure}
    
    \vspace{0.2cm}
    \begin{subfigure}{0.45\linewidth}
        \centering
        \includegraphics[width=\linewidth,trim={0 60 0 200},clip]{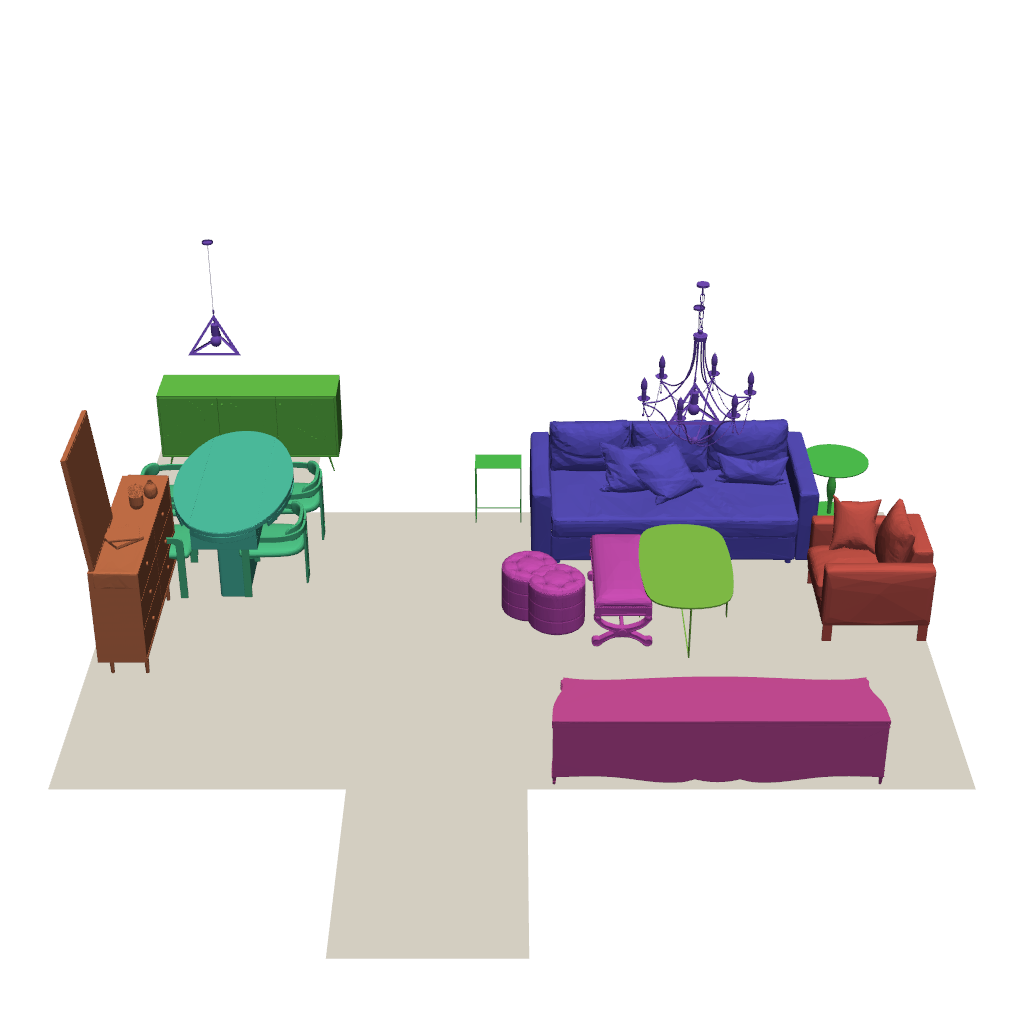}
    \end{subfigure}
    \hfill
    \begin{subfigure}{0.45\linewidth}
        \centering
        \includegraphics[width=\linewidth,trim={0 60 0 200},clip]{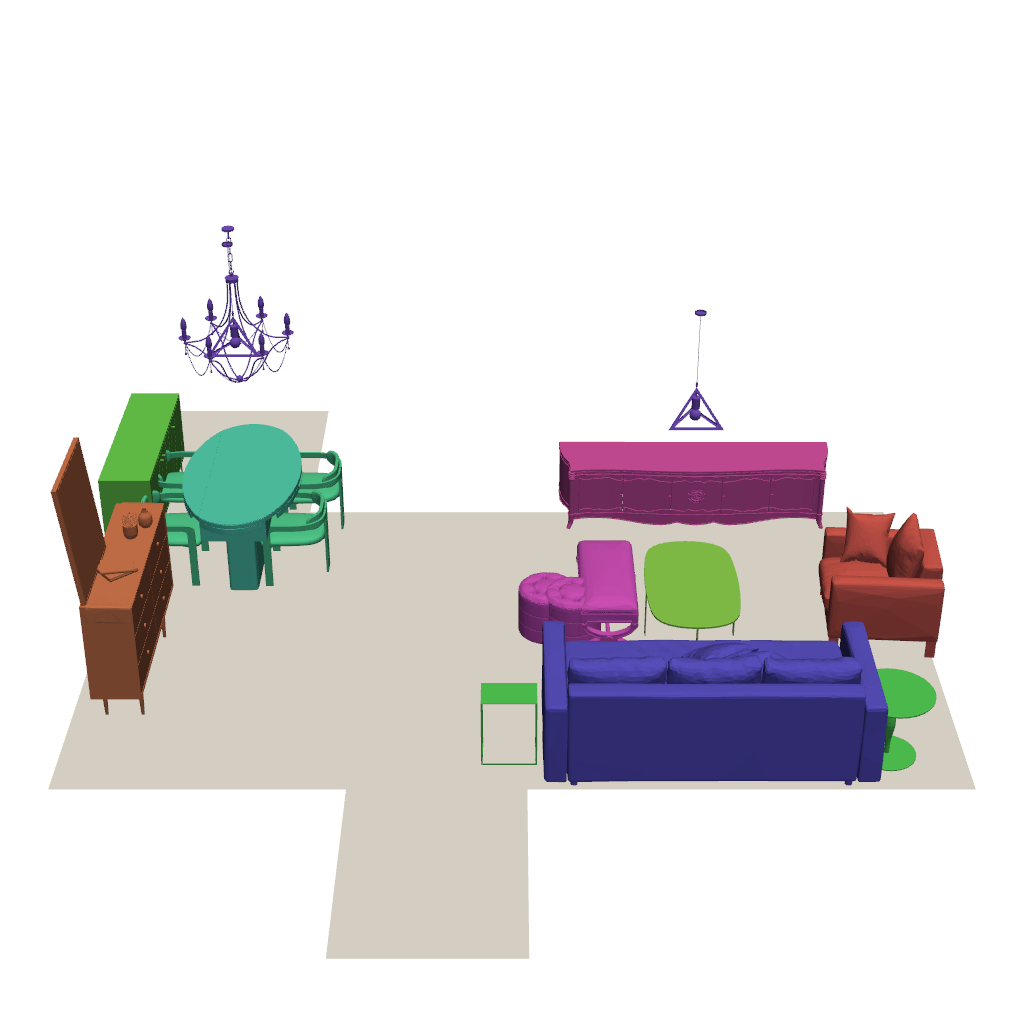}
    \end{subfigure}
    
    \caption{\textbf{Furniture arrangement.} Different table-chair arrangements (top) and different furniture placement directions (bottom).}
    \label{fig:furniture_arrangement}
\end{figure}
\algorithm can also be used for furniture arrangement, a scene synthesis problem conditioned on object labels and bounding box sizes without an initial configuration.
We use our proposed corruption-and-masking strategy over these attributes to test our models in this setting, and compare the results against DiffuScene's furniture arrangement variant (DiffuScene-FA). 
ATISS models are not suited for this task, since replacing the predicted object label by constraints breaks the stopping condition in their autoregressive pipeline.
Fig.~\ref{fig:furniture_arrangement} shows two pairs of example results with each pair generated from the same input.
We show quantitative results in Table~\ref{tab:furniture}. Without re-training, our models outperform DiffuScene-FA in all metrics.




\section{Limitations}

Though our method shows competitive performance in different layout generation tasks, it has several limitations: the current definition of objects as a collection of bounding box features and labels is not the most precise representation for a 3D setting, and gaining a good understanding of 3D features can potentially improve upon this aspect. In addition, \algorithm still requires a model retrieving strategy to compose the final 3D scene configuration. In the future we want to investigate the use of strong 3D shape priors (in the form of a point cloud or mesh encoder) and integrating mesh and texture synthesis capabilities such as SceneTex~\cite{Chen23arxiv-scenetex}, to enhance realism. 


\section{Conclusion}
In this work we introduced \algorithm, a novel mixed diffusion model combining DDPM and D3PM for object based 3D indoor scene synthesis. This formulation circumvents the need of adopting one-hot representations of semantic labels in the continuous domain as done in prior works.
We designed a denoising architecture that has a time-variant transformer decoder backbone and outputs predictions separately for discrete semantic attributes and continuous geometric attributes.
We proposed a corruption-and-masking strategy to conduct scene completion and furniture arrangement without the need to re-train our models.
We provided extensive experimental evaluation using common benchmarks against state-of-the-art autoregressive and diffusion models.
The results show a clear advantage of our approach over the state-of-the-art with or without partial constraints.
We also presented ablation studies supporting our design of the mixed diffusion framework as well as the denoising architecture.


{
    \small
    \bibliographystyle{ieeenat_fullname}

}

\clearpage
\appendix


\section{Diffusion Models\label{app:parameterization}}

\subsection{DDPM forward process re-parameterization, simplified loss}
DDPM~\cite{Ho20neurips-DiffusionModel} suggests re-parameterizing Eq.~\eqref{eq:forward_xt} as 
\begin{equation}
    \ContinuousVar_t(\ContinuousVar_0, \M{\epsilon}) =\sqrt{\Bar{\alpha}_t}\ContinuousVar_0 + \sqrt{1-\Bar{\alpha}_t}\M{\epsilon}
\end{equation} 
with $\M{\epsilon} \sim \calN(\M{0}, \MI)$.
Then, let $\M{\epsilon}_\theta$ be a function approximator to predict $\M{\epsilon}$ from $\ContinuousVar_t$ such that for the reverse process $\GaussianMean_\theta(\ContinuousVar_t, t) = \frac{1}{\sqrt{\alpha_t}} \left( \ContinuousVar_t - \frac{\beta_t}{\sqrt{1 - \Bar{\alpha}_t}} \M{\epsilon}_\theta(\ContinuousVar_t, t) \right)$ and $L_{t-1}$ for $2\leq t\leq T$ in ~\eqref{eq:diffusion_loss} becomes
\begin{equation}
\begin{split}
    L_{t-1} =& \E_{\ContinuousVar_0, \M{\epsilon}} \left[ \frac{\beta_t^2}{2\sigma_t^2 \alpha_t (1-\Bar{\alpha}_t)} \right. \\
    & \hspace{1cm} \left. \norm{\M{\epsilon} - \M{\epsilon}_\theta (\sqrt{\Bar{\alpha}_t} \ContinuousVar_0 + \sqrt{1-\Bar{\alpha}_t} \M{\epsilon}, t)}^2 \right] \\
    & + \text{const}    
\end{split}
\end{equation}
where the constant term can be dropped in training.

On image data, \cite{Ho20neurips-DiffusionModel} found it beneficial to sample quality to train on this simplified version of the variational bound ($L_{0:T-1}$):
\begin{equation}
    \label{eq:DDPM_loss}
    L_{\text{simple}}(\theta) := \E_{t, \ContinuousVar_0, \M{\epsilon}} \left[ \norm{\M{\epsilon} - \M{\epsilon}_\theta (\sqrt{\Bar{\alpha}_t} \ContinuousVar_0 + \sqrt{1-\Bar{\alpha}_t} \M{\epsilon}, t)}^2 \right]
\end{equation}
with uniform $t$ between 1 and $T$, dropping the scaling factors of the squared norms. We use Eq.~\eqref{eq:DDPM_loss} to compute $L_{vb}^{DDPM}$ in \algorithm.

\subsection{D3PM reverse parameterization, auxiliary loss}
Similar to DDPM, \cite{Austin21neurips-D3PM, Hoogeboom21neurips-MultinomialDiffusion} suggested that approximating some surrogate variables gives better quality. Specifically, they trained a neural network $\Tilde{p}_\networkVar(\Tilde{\DiscreteVar}_0| \DiscreteVar_t)$, multiplied it with the posterior $q(\DiscreteVar_{t-1} | \DiscreteVar_{t}, \DiscreteVar_0)$, and marginalized $\Tilde{\DiscreteVar}_0$ to obtain $p_\networkVar(\DiscreteVar_{t-1}| \DiscreteVar_t)$:
\begin{equation}
    p_\networkVar(\discreteVar_{t-1}| \discreteVar_t) \propto \sum_{\Tilde{\discreteVar}_0} q(\discreteVar_{t-1} | \discreteVar_{t}, \discreteVar_0) \Tilde{p}_\networkVar(\Tilde{\discreteVar}_0| \discreteVar_t).
\end{equation}

D3PM~\cite{Austin21neurips-D3PM} introduced an auxiliary denoising objective to encourage good predictions of the data $\DiscreteVar_0$ at each time step. Their complete training loss is:
\begin{equation}
    L_\lambda (\networkVar) = L_{vb} (\networkVar) + \lambda \E_{q(\discreteVar_0) q(\discreteVar_t | \discreteVar_0)}[-\log \Tilde{p}_\networkVar({\discreteVar}_0| \discreteVar_t)]
\end{equation}
which correspond to $L_{vb}^{D3PM}$ and $\lambda L_{aux}^{D3PM}$ in our \algorithm algorithm.
This can be implemented efficiently in training, since the parameterization of the backward step directly computes $\Tilde{p}_\networkVar({\DiscreteVar}_0| \DiscreteVar_t)$.

\section{\algorithm Loss Factorization \label{app:factorization}}
We provide more details for the factorization step in Eq.~\eqref{eq:kl_mixed}.
We first consider a general case comparing the KL-divergence between $\Tilde{q}(\discreteVar)\hat{q}(\ContinuousVar)$ and $\Tilde{p}(\discreteVar)\hat{p}(\ContinuousVar)$ for discrete $\discreteVar$ and continuous $\ContinuousVar$:
\begin{equation}
\label{eq:loss_factorization_general}
\begin{split}
    &D_{\mathrm{KL}}(\Tilde{q}(\discreteVar)\hat{q}(\ContinuousVar) || \Tilde{p}(\discreteVar) \hat{p}(\ContinuousVar))
    = \int_\ContinuousVar \sum_\discreteVar \Tilde{q}(\discreteVar)\hat{q}(\ContinuousVar) \log \frac{\Tilde{q}(\discreteVar)\hat{q}(\ContinuousVar)}{\Tilde{p}(\discreteVar) \hat{p}(\ContinuousVar)} \\
    &= \int_\ContinuousVar \sum_\discreteVar \left [
    \Tilde{q}(\discreteVar)\hat{q}(\ContinuousVar) \log \frac{\Tilde{q}(\discreteVar)}{\Tilde{p}(\discreteVar)} + \Tilde{q}(\discreteVar)\hat{q}(\ContinuousVar) \log \frac{\hat{q}(\ContinuousVar)}{\hat{p}(\ContinuousVar)} \right] \\
    &=   \sum_\discreteVar \Tilde{q}(\discreteVar) \log \frac{\Tilde{q}(\discreteVar)}{\Tilde{p}(\discreteVar)} \underbrace{\int_\ContinuousVar \hat{q}(\ContinuousVar)}_{1} + 
    \int_\ContinuousVar \hat{q}(\ContinuousVar) \log \frac{\hat{q}(\ContinuousVar)}{\hat{p}(\ContinuousVar)} \underbrace{\sum_\discreteVar \Tilde{q}(\discreteVar)}_{1} \\
    &= D_{\mathrm{KL}}(\Tilde{q}(\discreteVar) || \Tilde{p}(\discreteVar)) + D_{\mathrm{KL}}(\hat{q}(\ContinuousVar) || \Tilde{p}(\ContinuousVar)).
\end{split}
\end{equation}
This means we can decouple KL-divergence computation between mixed-domain probability distributions to a sum of domain-specific KL-divergence computation.
Note that this holds for any choices of $\Tilde{q}$, $\hat{q}$, $\Tilde{p}$, $\hat{p}$.
For our specific case:
\begin{equation}
\begin{split}
    &L_{t-1}^{mixed} =\mathbb{E}_{q(\discreteVar_t, \ContinuousVar_t | \discreteVar_0, \ContinuousVar_0)}
	\left [ 
	D_{\mathrm{KL}} (q(\discreteVar_{t-1}, \ContinuousVar_{t-1} | \discreteVar_t, \ContinuousVar_t, \discreteVar_0, \ContinuousVar_0) \right.\\
        & \hspace{4.5cm} \left. || p_\theta (\discreteVar_{t-1}, \ContinuousVar_{t-1} | \discreteVar_t, \ContinuousVar_t, \CondVar) )
	\right] \\  
	&= \mathbb{E}_{\Tilde{q}(\discreteVar_t|\discreteVar_0) \hat{q}(\ContinuousVar_t|\ContinuousVar_0)}
	\left [D_{\mathrm{KL}} ( \Tilde{q}(\discreteVar_{t-1} | \discreteVar_t,  \discreteVar_0) \hat{q}(\ContinuousVar_{t-1} | \ContinuousVar_t,  \ContinuousVar_0) \right.\\
	& \hspace{3cm} \left. || \Tilde{p}_\theta (\discreteVar_{t-1} | \discreteVar_t, \ContinuousVar_t, \CondVar) ) \hat{p}_\theta (\ContinuousVar_{t-1} | \ContinuousVar_t, \discreteVar_t, \CondVar) )
	\right] \\
	&= \mathbb{E}_{\Tilde{q}(\discreteVar_t|\discreteVar_0)} \left[
	D_{\mathrm{KL}} ( \Tilde{q}(\discreteVar_{t-1} | \discreteVar_t,  \discreteVar_0) || \Tilde{p}_\theta (\discreteVar_{t-1} | \discreteVar_t, \ContinuousVar_t, \CondVar) ))
	\right] \\
	& \quad + \mathbb{E}_{\hat{q}(\ContinuousVar_t|\ContinuousVar_0)} \left[
	D_{\mathrm{KL}} ( \hat{q}(\ContinuousVar_{t-1} | \ContinuousVar_t,  \ContinuousVar_0)  || \hat{p}_\theta (\ContinuousVar_{t-1} | \ContinuousVar_t, \discreteVar_t, \CondVar))
	\right],
\end{split}
\end{equation}
where the last equality holds because of Eq.~\eqref{eq:loss_factorization_general}.

\section{Network and Implementation Details \label{app:implementation}}
We include detailed network hyper-parameters for our denoising network, and implementation for training. 

\subsection{Network Hyper-parameters}
The architecture of MiDiffusion has at its core a series of 8 transformer blocks using a hidden dimension of 512 with 4 heads and feed-forward layers with an internal dimension of 2048. Following~\cite{Gu22cvpr-VQ-Diffusion}, we use GELU~\cite{Hendrycks16arxiv-gelu} as nonlinearity. 
The semantic attributes are embedded in a learned vector of length 512, and the geometric attributes are mapped by a 3-layer MLP, with internal dimensions of [512, 1024], to a 512-dimensional space before being fed to the transformer decoder blocks. 
We extract floor plan features of dimension 64 using a 4-layer PointNet~\cite{Qi17cvpr-pointnet} with internal dimensions [64, 64, 512] as in LEGO-Net~\cite{Wei23cvpr-LEGONet}. 
For baselines and ablation studies, we use the default ResNet-18~\cite{He16cvpr-ResNet} image feature extractor proposed by ATISS~\cite{Paschalidou21neurips-ATISS} and also implemented by DiffuScene~\cite{Tang24cvpr-DiffuScene} to compute the 64-dimensional floor plan features from binary floor plan masks.
The outputs of the transformer decoder are fed to two separate MLPs to decode the semantic and geometric predictions. The semantic feature decoder is a 1-layer MLP that produces a categorical distribution over $\discreteVar$, and the geometric feature decoder is a 3-layer MLP with hidden dimensions [512, 1024], producing the 8-dimensional Gaussian mean for the geometric attributes $\ContinuousVar$.

\subsection{Implementation}
We set a fixed learning rate of $l_r=2e^{-4}$ using the Adam optimizer, and a dropout ratio of 0.1 for multi-head attention and feed-forward layers in the transformer blocks.
We train 50k epochs on the bedroom dataset with $0.5$ learning rate decay every $10k$ epochs, and $100k$ epochs on the living and dining room datasets with $0.5$ decay every $15k$ epochs.
We use a linear schedule over 1000 diffusion steps for all noise parameters in the forward process. In the discrete domain $\alpha_t$ and $\gamma_t$ range from $1-1e^{-5}$ to $0.99999$ and from $9e^{-6}$ to $0.99999$ respectively. In the continuous domain, $\beta_t$ starts from $1e^{-4}$ at reaches $0.02$. 
We train all our models on a single NVIDIA V100 GPU with under 8GB of GPU RAM usage. The training time range from around 20 hours on living and dining room datasets to about 32 hours on the bedroom dataset for a batch size of 512 scenes.



\begin{table*}[!ht]
\resizebox{\textwidth}{!}{
\begin{tabular}{llcccclcccclcccc}
\hline
\multirow{2}{*}{Method} &  & \multicolumn{4}{c}{Bedroom}  &  & \multicolumn{4}{c}{Dining room} &  & \multicolumn{4}{c}{Living room} \\ \cline{3-6} \cline{8-11} \cline{13-16} 
               &  & FID   &KIDx0.001& CA \%   &KLx0.01&  & FID   &KIDx0.001& CA \%   &\small{KLx0.01}&  & FID   &KIDx0.001& CA \%   &KLx0.01\\ \hline

\multicolumn{2}{l}{2-stage + PointNet}  & 65.54	&  1.53	  & 57.83 $\pm$ 4.89   & 3.50      &  & 38.80	& 7.02         & 61.53 $\pm$ 7.85      &    5.08    &  & 35.35	  & 6.30		  & 61.97 $\pm$ 6.55      & 4.73       \\
\multicolumn{2}{l}{2-stage (gt labels) + PointNet}  & 64.37	 &  0.67   & 56.19 $\pm$ 5.45          & --          &  & 32.47	  & 5.07	    & 56.29 $\pm$ 4.80         & --          &  & 32.26  & 5.86		       & 58.85 $\pm$ 6.29      & --          \\ 
\hline
\multicolumn{2}{l}{\textbf{Mixed+PointNet}}  & \textbf{63.71}        & \textbf{0.24}     & \textbf{53.65 $\pm$ 2.46} & \textbf{1.05} & & \textbf{30.63} & \textbf{2.74} & \textbf{52.10 $\pm$ 1.72} & \textbf{1.90} &         & \textbf{28.60} & \textbf{1.64} & \textbf{54.04 $\pm$ 3.33}  & \textbf{1.54} \\
 \hline
\end{tabular}
}
\vspace{-0.2cm}
\caption{Additional ablation study for floor-conditioned 3D scene synthesis. \label{tab:conditional_ablation_app}} 
\end{table*}



\begin{table*}[!htbp]
\centering
\resizebox{0.7\textwidth}{!}{
\begin{tabular}{llccclccclccc}
\hline
\multirow{2}{*}{Method} &  & \multicolumn{3}{c}{Bedroom}  &  & \multicolumn{3}{c}{Dining room} &  & \multicolumn{3}{c}{Living room} \\ \cline{3-5} \cline{7-9} \cline{11-13} 
            &   & Obj           & OOB \%        & IoU \%         &           & Obj            & OOB \%         & IoU \%         & \textbf{} & Obj            & OOB \%        & IoU \%         \\ \hline
2stage + PointNet &   & 5.04	     & 5.67         & 0.72        &  & \textbf{11.30}	& 11.80  & 	1.57          &  & 11.50 &	13.35	& 1.48         \\
2stage (gt labels) + PointNet &   & 	 --    & 5.20         & 0.61        &  & --	& 8.87  & 	1.18          &  & -- &	11.80	& 1.28         \\ \hline
DDPM+ResNet       &   & 5.07          & 6.63         & 0.50        &  & 10.87          & 7.08          & \textbf{0.70}          &  & 11.90          & 10.93         & 0.70          \\
\Update{Mixed+ResNet}{}&   & \Update{5.15}{}      & \Update{4.43}{}         & \Update{\textbf{0.44}  }{}      &  & \Update{10.78}{}          & \Update{7.09}{}          & \Update{0.73}{}         &  & \Update{11.91}{}    & \Update{10.60}{}         & \Update{0.70}{}          \\
DDPM+PointNet  &   & 4.92          & 5.29          & 0.70 &  & 10.78          & 6.75         & 0.78 &  & \textbf{11.83} & 10.37         & \textbf{ 0.64} \\ \hline
\textbf{Mixed+PointNet}       &   & \textbf{5.22} & \textbf{3.91} & 0.61         &  & \textbf{10.92} & \textbf{5.77}  & 0.91          &  & 11.91          & \textbf{7.93} & 0.98         \\ \hline
Ground truth          &   & 5.22          & 3.37          & 0.24          &  & 11.11          & 0.73          & 0.48          &  & 11.67          & 1.55          & 0.27  \\ \hline

\end{tabular}
}
\vspace{-0.2cm}
\caption{Geometric evaluations on ablation study for floor-conditioned 3D scene synthesis. \label{tab:ablation_obj}} 
\end{table*}


\begin{table*}[!htbp]
\resizebox{\textwidth}{!}{
\begin{tabular}{llcccclcccclcccc}
\hline
\multirow{2}{*}{Method} &  & \multicolumn{4}{c}{Bedroom}  &  & \multicolumn{4}{c}{Dining room} &  & \multicolumn{4}{c}{Living room} \\ \cline{3-6} \cline{8-11} \cline{13-16} 
            &  & FID   &KIDx0.001& OBB \%   &IoU \% &  & FID   &KIDx0.001& OBB \%   &IoU \% &  & FID   &KIDx0.001& OBB \%   &IoU \% \\ \hline
\multicolumn{2}{l}{ {MiDiffusion + DM}}  &  {59.43}   &  {-0.07}     &  {7.71} &  {0.56} & &  {32.17} &  {4.46} &  {6.49} &  {\textbf{0.95}} &         &  {30.47} &  {3.65} &  {8.72} &  {\textbf{0.84}} \\
\multicolumn{2}{l}{MiDiffusion (ours)}  & \textbf{59.08}   & \textbf{-0.24}     & \textbf{7.34} & \textbf{0.55} & & \textbf{31.72} & \textbf{3.91} & \textbf{6.35} & 0.96 &         & \textbf{29.79} & \textbf{3.13} & \textbf{8.33} & \textbf{0.84} \\
 \hline
\end{tabular}
}
\vspace{-0.2cm}
\caption{Additional evaluation results for furniture arrangement experiment.  \label{tab:furniture_masking}} 
\end{table*}

\section{Example Synthesized Layout Images \label{app:layout_synthesis}}

\begin{figure}[ht]
    \centering
    \begin{minipage}[t]{0.12\linewidth}
       \tiny{ATISS\\~\cite{Paschalidou21neurips-ATISS}}
    \end{minipage}
    \begin{minipage}{.28\linewidth}
        \centering
        \includegraphics[width=0.98\linewidth]{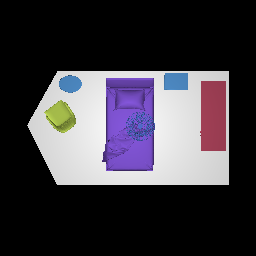}
    \end{minipage}%
    \begin{minipage}{0.28\linewidth}
        \centering
        \includegraphics[width=0.98\linewidth]{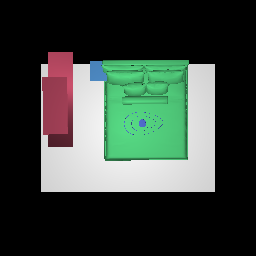}
    \end{minipage}
    \begin{minipage}{.28\linewidth}
        \centering
        \includegraphics[width=0.98\linewidth]{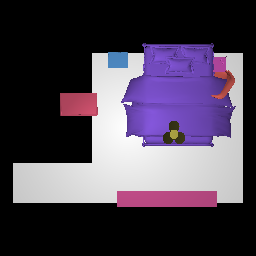}
    \end{minipage}%

    \centering
    \begin{minipage}[t]{0.12\linewidth}
       \tiny{DiffuScene\\~\cite{Tang24cvpr-DiffuScene}}
    \end{minipage}
    \begin{minipage}{.28\linewidth}
        \centering
        \includegraphics[width=0.98\linewidth]{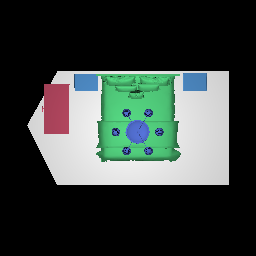}
    \end{minipage}%
    \begin{minipage}{0.28\linewidth}
        \centering
        \includegraphics[width=0.98\linewidth]{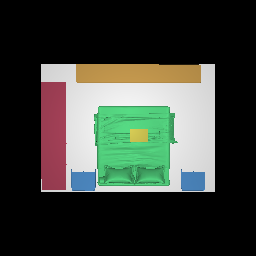}
    \end{minipage}
    \begin{minipage}{.28\linewidth}
        \centering
        \includegraphics[width=0.98\linewidth]{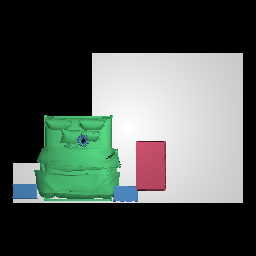}
    \end{minipage}%
    
    \centering
    \begin{minipage}[t]{0.12\linewidth}
        \tiny{\algorithm\\(Ours)}
    \end{minipage}
    \begin{minipage}{.28\linewidth}
        \centering
        \includegraphics[width=0.98\linewidth]{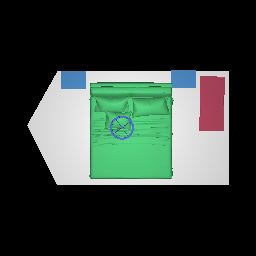}
    \end{minipage}%
    \begin{minipage}{0.28\linewidth}
        \centering
        \includegraphics[width=0.98\linewidth]{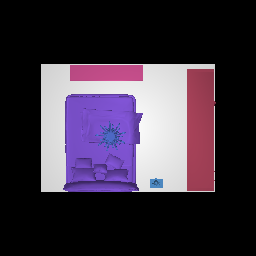}
    \end{minipage}
    \begin{minipage}{.28\linewidth}
        \centering
        \includegraphics[width=0.98\linewidth]{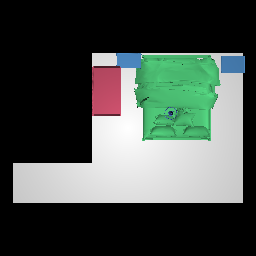}
    \end{minipage}%

    \centering
    \begin{minipage}[t]{0.12\linewidth}
        \tiny{Ground-truth}
    \end{minipage}
    \begin{minipage}{.28\linewidth}
        \centering
        \includegraphics[width=0.98\linewidth]{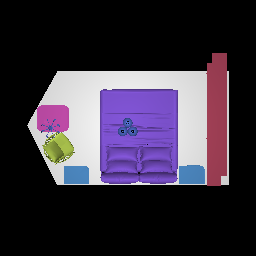}
    \end{minipage}%
    \begin{minipage}{0.28\linewidth}
        \centering
        \includegraphics[width=0.98\linewidth]{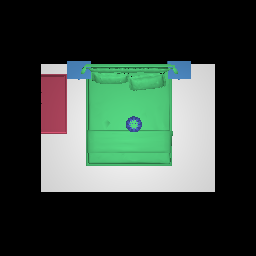}
    \end{minipage}
    \begin{minipage}{.28\linewidth}
        \centering
        \includegraphics[width=0.98\linewidth]{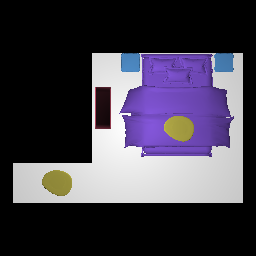}
    \end{minipage}%
    \caption{Example bedroom top-down orthographic projection images for quantitative evaluations.\label{fig:bedroom_rendered_layouts}}
\end{figure}

\begin{figure}[ht]
    \centering
    \begin{minipage}[t]{0.12\linewidth}
       \tiny{ATISS\\~\cite{Paschalidou21neurips-ATISS}}
    \end{minipage}
    \begin{minipage}{.28\linewidth}
        \centering
        \includegraphics[width=0.98\linewidth]{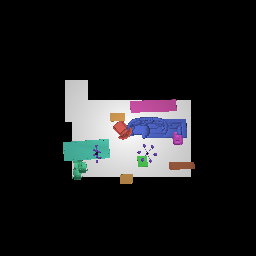}
    \end{minipage}%
    \begin{minipage}{0.28\linewidth}
        \centering
        \includegraphics[width=0.98\linewidth]{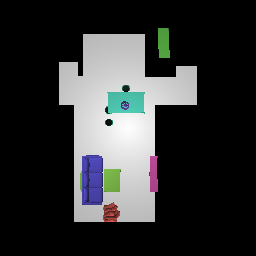}
    \end{minipage}
    \begin{minipage}{.28\linewidth}
        \centering
        \includegraphics[width=0.98\linewidth]{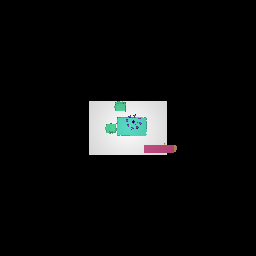}
    \end{minipage}%

    \centering
    \begin{minipage}[t]{0.12\linewidth}
       \tiny{DiffuScene\\~\cite{Tang24cvpr-DiffuScene}}
    \end{minipage}
    \begin{minipage}{.28\linewidth}
        \centering
        \includegraphics[width=0.98\linewidth]{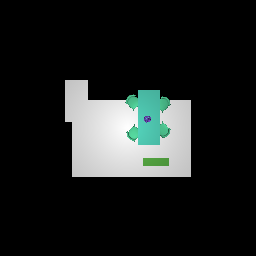}
    \end{minipage}%
    \begin{minipage}{0.28\linewidth}
        \centering
        \includegraphics[width=0.98\linewidth]{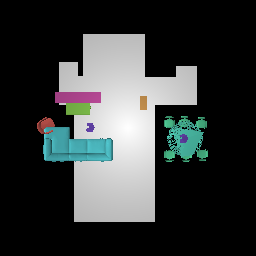}
    \end{minipage}
    \begin{minipage}{.28\linewidth}
        \centering
        \includegraphics[width=0.98\linewidth]{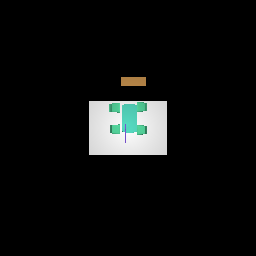}
    \end{minipage}%
    
    \centering
    \begin{minipage}[t]{0.12\linewidth}
       \tiny{\algorithm\\(Ours)}
    \end{minipage}
    \begin{minipage}{.28\linewidth}
        \centering
        \includegraphics[width=0.98\linewidth]{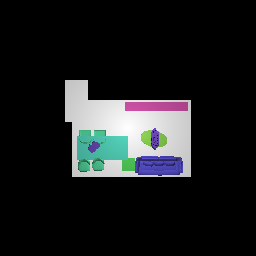}
    \end{minipage}%
    \begin{minipage}{0.28\linewidth}
        \centering
        \includegraphics[width=0.98\linewidth]{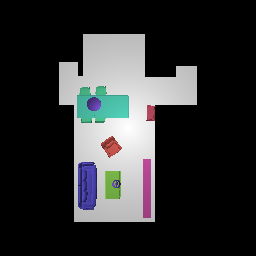}
    \end{minipage}
    \begin{minipage}{.28\linewidth}
        \centering
        \includegraphics[width=0.98\linewidth]{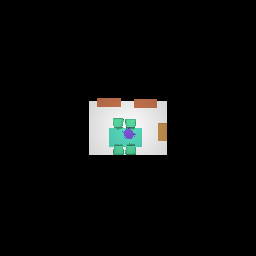}
    \end{minipage}%

    \centering
    \begin{minipage}[t]{0.12\linewidth}
       \tiny{Ground-truth}
    \end{minipage}
    \begin{minipage}{.28\linewidth}
        \centering
        \includegraphics[width=0.98\linewidth]{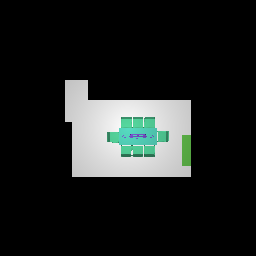}
    \end{minipage}%
    \begin{minipage}{0.28\linewidth}
        \centering
        \includegraphics[width=0.98\linewidth]{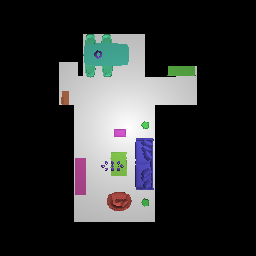}
    \end{minipage}
    \begin{minipage}{.28\linewidth}
        \centering
        \includegraphics[width=0.98\linewidth]{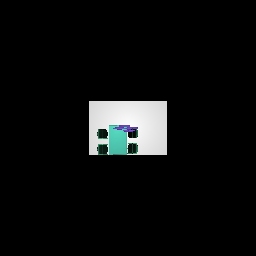}
    \end{minipage}%
    
    \caption{Example dining room top-down orthographic projection images for quantitative evaluations. \label{fig:diningroom_rendered_layouts}}
    
\end{figure}

\begin{figure}[ht]
    \centering
    \begin{minipage}[t]{0.12\linewidth}
       \tiny{ATISS\\~\cite{Paschalidou21neurips-ATISS}}
    \end{minipage}
    \begin{minipage}{.28\linewidth}
        \centering
        \includegraphics[width=0.98\linewidth]{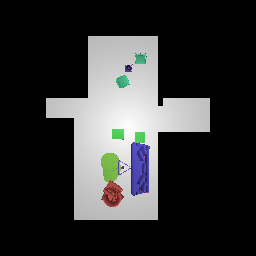}
    \end{minipage}%
    \begin{minipage}{0.28\linewidth}
        \centering
        \includegraphics[width=0.98\linewidth]{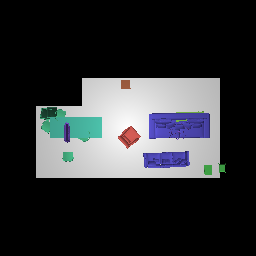}
    \end{minipage}
    \begin{minipage}{0.28\linewidth}
        \centering
        \includegraphics[width=0.98\linewidth]{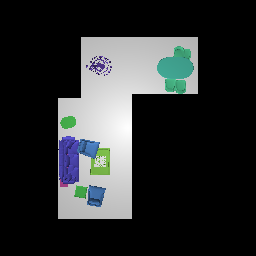}
    \end{minipage}%

    \centering
    \begin{minipage}[t]{0.12\linewidth}
       \tiny{DiffuScene\\~\cite{Tang24cvpr-DiffuScene}}
    \end{minipage}
    \begin{minipage}{0.28\linewidth}
        \centering
        \includegraphics[width=0.98\linewidth]{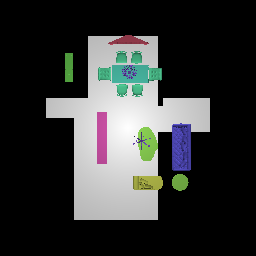}
    \end{minipage}%
    \begin{minipage}{0.28\linewidth}
        \centering
        \includegraphics[width=0.98\linewidth]{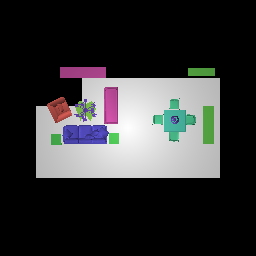}
    \end{minipage}
    \begin{minipage}{0.28\linewidth}
        \centering
        \includegraphics[width=0.98\linewidth]{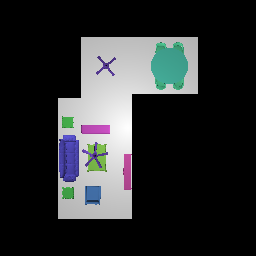}
    \end{minipage}%
    
    \centering
    \begin{minipage}[t]{0.12\linewidth}
        \tiny{\algorithm\\(Ours)}
    \end{minipage}
    \begin{minipage}{0.28\linewidth}
        \centering
        \includegraphics[width=0.98\linewidth]{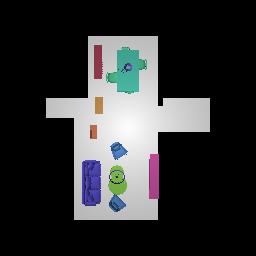}
    \end{minipage}%
    \begin{minipage}{0.28\linewidth}
        \centering
        \includegraphics[width=0.98\linewidth]{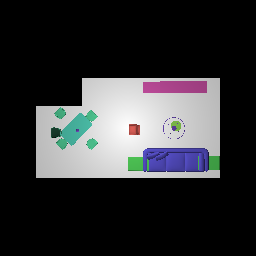}
    \end{minipage}
    \begin{minipage}{0.28\linewidth}
        \centering
        \includegraphics[width=0.98\linewidth]{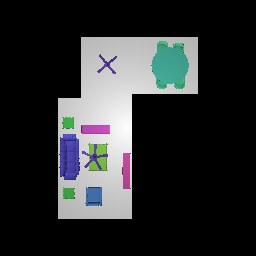}
    \end{minipage}%

    \centering
    \begin{minipage}[t]{0.12\linewidth}
        \tiny{Ground-truth}
    \end{minipage}
    \begin{minipage}{0.28\linewidth}
        \centering
        \includegraphics[width=0.98\linewidth]{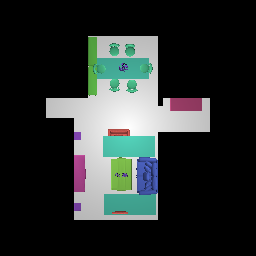}
    \end{minipage}%
    \begin{minipage}{0.28\linewidth}
        \centering
        \includegraphics[width=0.98\linewidth]{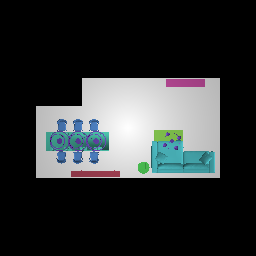}
    \end{minipage}
    \begin{minipage}{0.28\linewidth}
        \centering
        \includegraphics[width=0.98\linewidth]{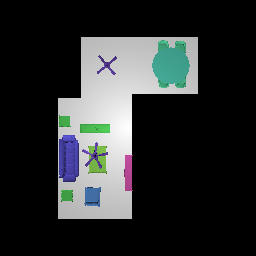}
    \end{minipage}%
    
    \caption{Example living room top-down orthographic projection images for quantitative evaluations. \label{fig:livingroom_rendered_layouts}}
\end{figure}

We use texture-less objects rendered on white floor plans for quantitative evaluations, since floor and object textures affect the clarity of the results. 
The object labels are sorted in alphabetical order for each room type, and then the associated colors are evenly sampled along a circular path using Seaborn's~\cite{Waskom21-seaborn} ``hls'' color palette.
We render top-down layout images using simple-3dviz~\cite{Katharopoulos20-simple3dviz} in accordance with prior works~\cite{Paschalidou21neurips-ATISS, Tang24cvpr-DiffuScene}.
We include examples of rendered images for three comparing approaches in Fig.~\ref{fig:bedroom_rendered_layouts},~\ref{fig:diningroom_rendered_layouts},~\ref{fig:livingroom_rendered_layouts} from the floor-conditioned scene synthesis experiment in Sec~\ref{fig:floor_conditioned}. 
Some ground-truth layouts include objects slightly out of boundary. This is a problem in the raw 3D-FRONT dataset. Therefore, we inflate the room boundary by 0.1m when counting the number of out of boundary objects.
The bedroom datasets is the easiest (less number of objects, smaller in room size, more training data) so that all comparing approaches can generate good predictions.
On the harder living room and dining room datasets, \algorithm clearly outperforms the baselines by generating realistic geometric arrangements with desired symmetry and alignment between objects,  while respecting the floor boundary constraints. DiffuScene is also capable of generating good geometric arrangements but their architecture is less optimal for learning floor boundary constraints.



\section{Additional Experimental Evaluations \label{app:add_experiment}}
We provide supplementary results to complement Sec.~\ref{sec:experiment}. 
We begin with additional results from the ablation studies to further support our proposed mixed diffusion formulation and network design.
Then, we present more results on scene synthesis with partial constraints, including a comparison with the direct masking strategy and applications to other conditional setups.
Lastly, we show quantitative comparison between \algorithm and DiffuScene in the unconditioned scene synthesis experiment, which is original target application of DiffuScene.

\subsection{Ablation Study \label{app:ablation}}
We show an additional ablation study comparing \algorithm (also referred as Mixed+PointNet in ablation studies) and the 2-stage approach to supplement Table \ref{tab:conditional_ablation}. The second row of Table~\ref{tab:conditional_ablation_app} reports results of a 2-stage variant where ground-truth semantic labels are sent to the geometric diffusion stage. 
While bypassing the first semantic diffusion stage improves performance, this variant still under-performs compared to \algorithm. This suggests that even with a perfect first-stage label predictor, the 2-stage approach still produces less realistic scene layouts compared to ours.

We show object evaluation results in Table~\ref{tab:ablation_obj}.
Across all three datasets, there is a consistent improvement over OOB \% as we modify the floor plan feature extractor and diffusion framework towards our final design of \algorithm. This is consistent with the main results in Table~\ref{tab:conditional_ablation}.

\subsection{Masking Comparison\label{app:masking}}
In Sec.~\ref{sec:exp_object_condition}, we show that, with the proposed corruption-and-masking strategy, \algorithm models are able to generate room layouts under various types of object constraints. Compared to ours, direct masking at each denoising step is less principled. We provide an example comparison in Table~\ref{tab:furniture_masking} to demonstrate that our proposed masking strategy yields better results.

\subsection{Additional Applications \label{app:applications}}
\begin{figure}[!ht]
    \centering
    \begin{subfigure}{.4\linewidth}
        \includegraphics[width=\linewidth,trim={80 0 80 80},clip]{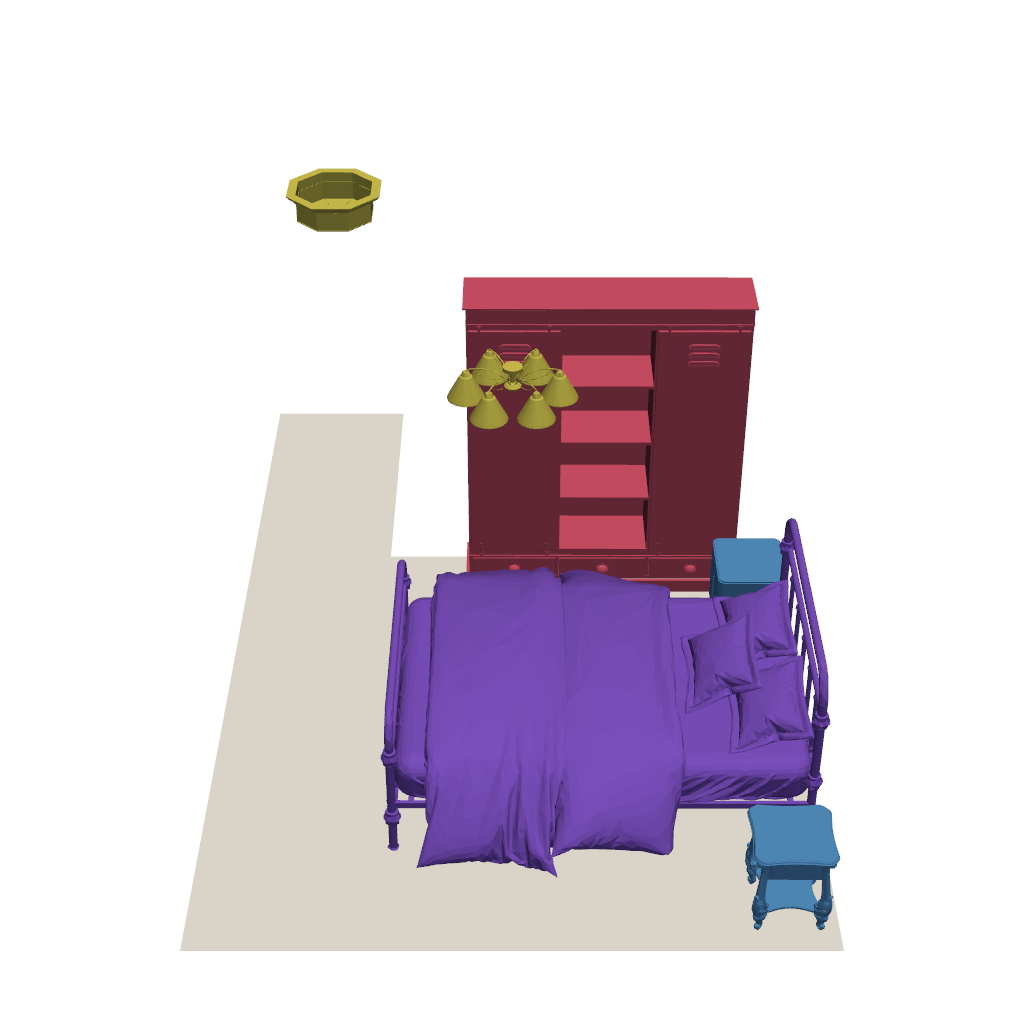}
    \end{subfigure}
    \hfill
    \begin{subfigure}{.45\linewidth}
        \includegraphics[width=\linewidth,trim={20 0 100 240},clip]{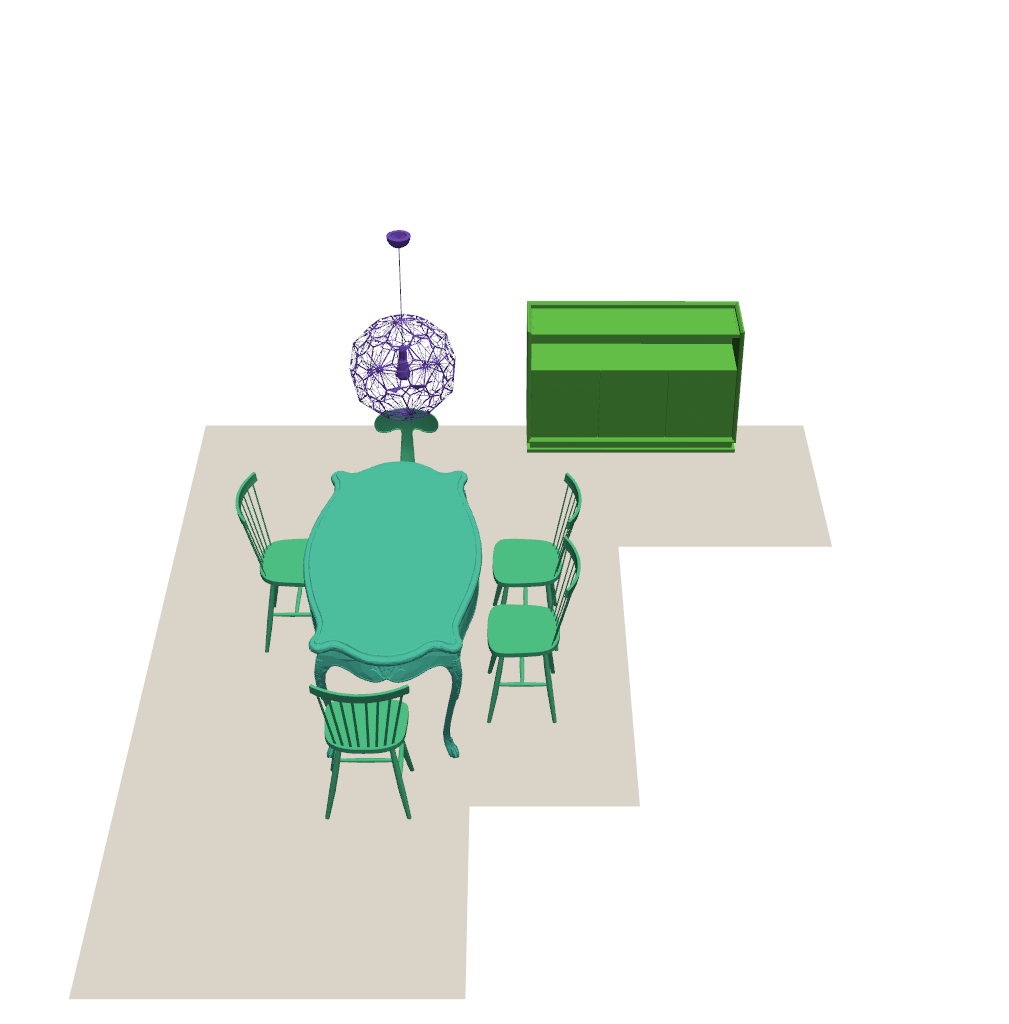}
    \end{subfigure}
    \caption{\textbf{Label constrained scene synthesis.} Bedroom (left) and dining room (bottom) examples. \label{fig:label_constrainted}}
\end{figure}

\begin{figure}[ht]
    \centering
    \begin{subfigure}{0.4\linewidth}
        \centering
        \includegraphics[width=\linewidth,trim={80 60 160 60},clip]{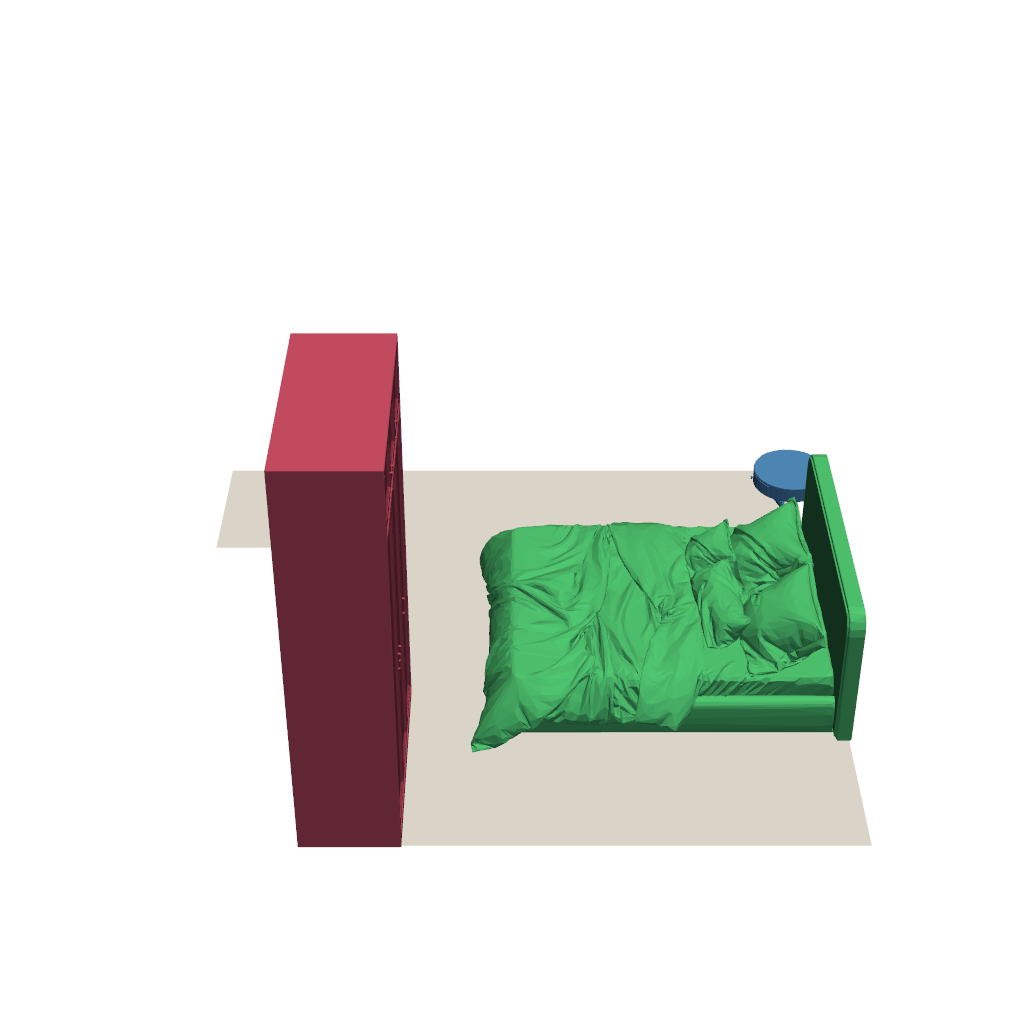}
    \end{subfigure}
    \hfill
    \begin{subfigure}{0.4\linewidth}
        \centering
        \includegraphics[width=\linewidth,trim={160 60 80 60},clip]{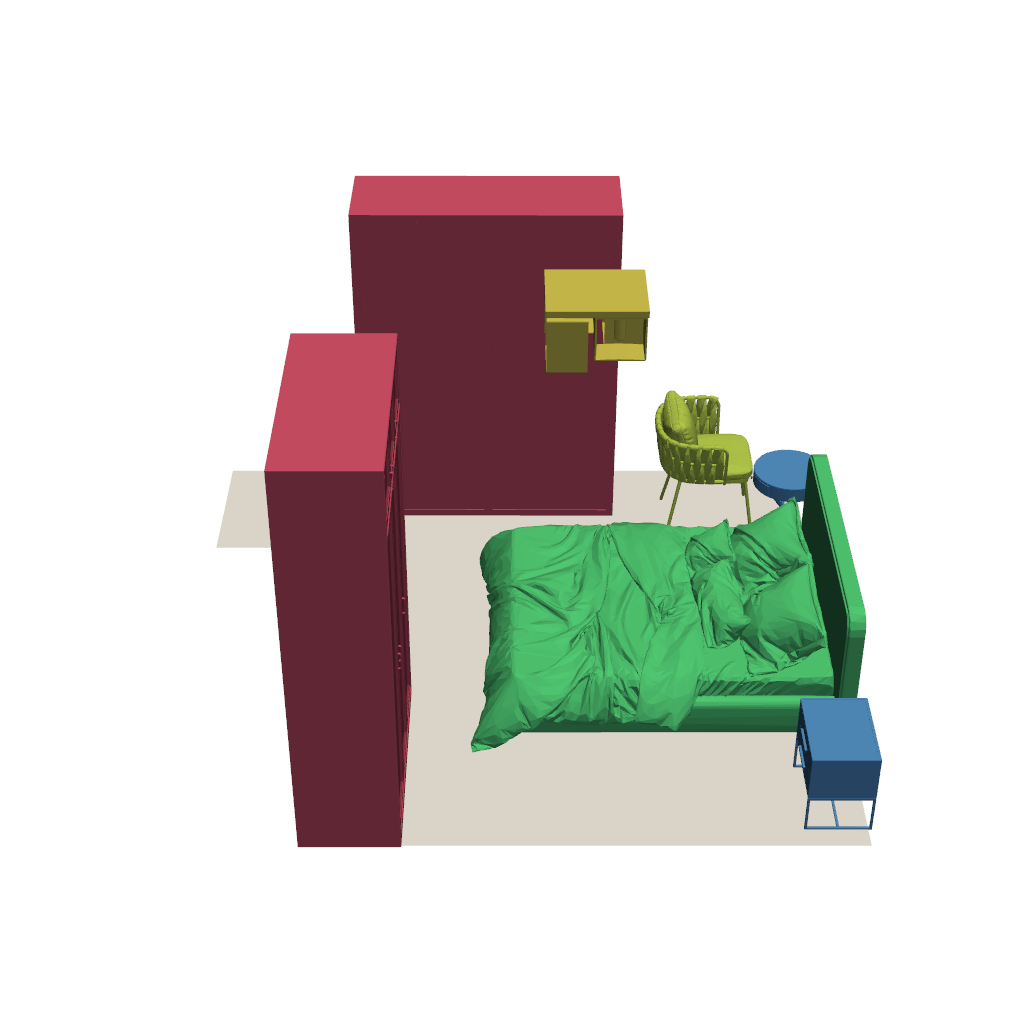}
    \end{subfigure}
    
    \begin{subfigure}{0.45\linewidth}
        \centering
        \includegraphics[width=\linewidth,trim={200 0 160 200},clip]{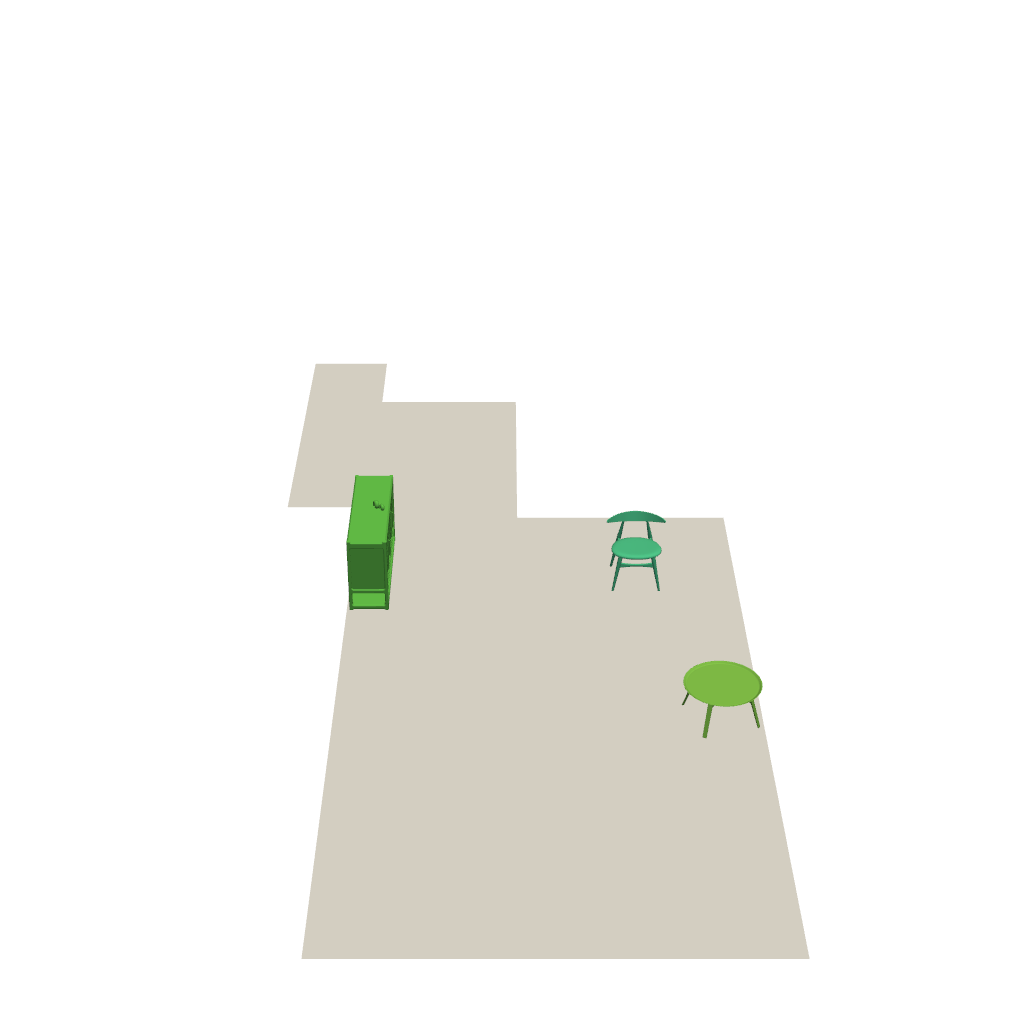}
    \end{subfigure}
    \hfill
    \begin{subfigure}{0.45\linewidth}
        \centering
        \includegraphics[width=\linewidth,trim={200 0 160 200},clip]{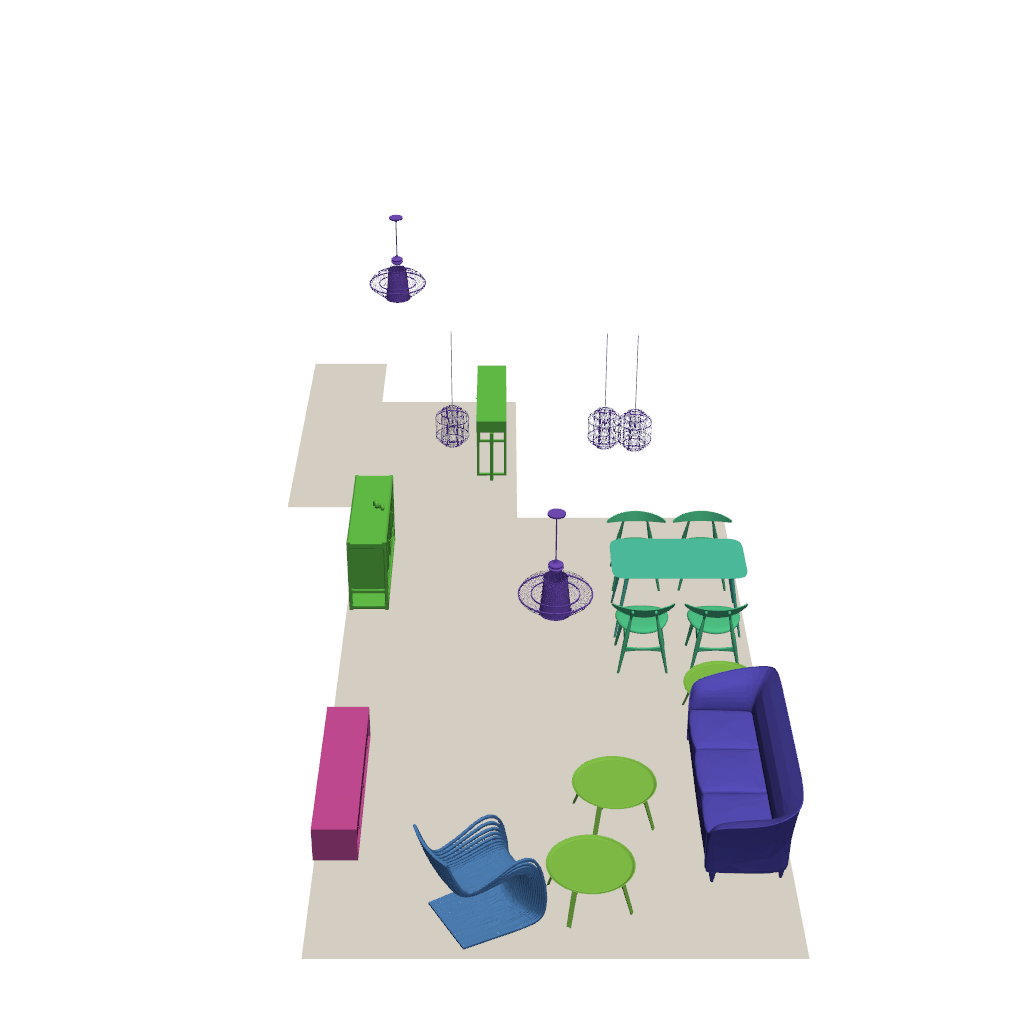}
    \end{subfigure}
    \caption{\textbf{Scene completion with label constraints}. Bedroom (top) and living room (bottom) label constrained scene completion examples. \label{fig:sc_label_constrainted}}
\end{figure}

Our models can easily extend to other applications with different masking ranges. For example, they can be applied to object category constrained scene synthesis given a set of desired object labels.  
Fig.~\ref{fig:label_constrainted} shows two examples of label constrained scene synthesis. In particular, the dining room scene is conditioned on five dining chairs, which occur less often than an even number of chairs. Our model is still able to generate reasonable layouts. 
We can even combine partial constraints on object attributes (\eg label) with existing objects. Fig.~\ref{fig:sc_label_constrainted} shows examples of such label constrained scene completion problems.

\subsection{Unconditional 3D Scene Synthesis\label{app:unconditional}}



\begin{table*}[!ht]
\resizebox{\textwidth}{!}{
\begin{tabular}{llcccclcccclcccc}
\hline
\multirow{2}{*}{Method} &  & \multicolumn{4}{c}{Bedroom}  &  & \multicolumn{4}{c}{Dining room} &  & \multicolumn{4}{c}{Living room} \\ \cline{3-6} \cline{8-11} \cline{13-16} 
            &  & FID   &KIDx0.001& CA \%   &KLx0.01&  & FID   &KIDx0.001& CA \%   &\small{KLx0.01}&  & FID   &KIDx0.001& CA \%   &KLx0.01\\ \hline
DiffuScene \cite{Tang24cvpr-DiffuScene}  &  & \textbf{61.12} & \textbf{0.46} & \textbf{51.09 $\pm$ 0.47} & 1.24 &  & 45.04         & 0.70 & 51.53 $\pm$ 1.05          & 1.10          &  & 43.30          & \textbf{0.21}          & 53.50 $\pm$ 2.84          & 0.67          \\
\multicolumn{2}{l}{MiDiffusion (ours)}  & 62.02 &	1.12	 & 52.36 $\pm$ 1.54	& \textbf{0.95} &  & \textbf{43.37}	& \textbf{0.68}	& \textbf{51.21 $\pm$ 0.82}	& \textbf{0.70} &  & \textbf{42.43} &	0.77 &	\textbf{53.46 $\pm$ 1.93} &	\textbf{0.58} \\
 \hline
\end{tabular}
}
\vspace{-0.2cm}
\caption{Evaluation results for unconditional 3D scene synthesis. \label{tab:unconditional}} 
\end{table*}



\begin{table*}[!ht]
\resizebox{\textwidth}{!}{
\begin{tabular}{llcccclcccclcccc}
\hline
\multirow{2}{*}{Method} &  & \multicolumn{4}{c}{Bedroom}  &  & \multicolumn{4}{c}{Dining room} &  & \multicolumn{4}{c}{Living room} \\ \cline{3-6} \cline{8-11} \cline{13-16} 
            &  & FID   &KIDx0.001& CA \%   &KLx0.01&  & FID   &KIDx0.001& CA \%   &{KLx0.01}&  & FID   &KIDx0.001& CA \%   &KLx0.01\\ \hline
DiffuScene (from \cite{Tang24cvpr-DiffuScene})      &  & 17.21         & 0.70          & 52.15           & 0.35          &  & 32.60          & 0.72         & 55.50          & 0.22          &  & 36.18          & 0.88         & 57.81          & 0.21          \\ \hline
DiffuScene \cite{Tang24cvpr-DiffuScene}  &  & \textbf{17.43} & \textbf{0.82} & \textbf{51.09 $\pm$ 0.47} & 0.66 &  & 33.07          & \textbf{0.93} & 53.82 $\pm$ 4.01          & 0.34          &  & {35.27}          & \textbf{0.58}          & {54.24 $\pm$ 3.65}          & 0.36          \\
\multicolumn{2}{l}{\algorithm(Ours)}      & 18.17	& \textbf{0.82}	& 52.36 $\pm$ 1.54 &	\textbf{0.16} &  & \textbf{31.39} &	1.15 &	\textbf{52.38 $\pm$ 1.85} &	\textbf{0.18} &  & \textbf{32.49}	& 1.20 &	\textbf{51.86 $\pm$ 1.80} &	\textbf{0.15} \\
 \hline
\end{tabular}
}
\vspace{-0.2cm}
\caption{Evaluation results for unconditional 3D scene synthesis against training scenes and using layouts projected to 6.2 squares. \label{tab:unconditional_additional}} 
\end{table*}

DiffuScene was originally proposed for 3D scene synthesis without floor conditioning. Although this is not the main focus of this paper, we provide experimental results comparing \algorithm and DiffuScene for completeness using the released model weights. Since this is an unconditioned problem, we slightly modify the proposed architecture in Fig.~\ref{fig:architecture} by removing modules related to the conditional input, including the floor plan feature encoder, the second AdaLN and the second multi-head cross-attention unit in each transformer block. We re-use all other hyper-parameters from the floor-conditioned experiments.

We follow the same layout image generation and evaluation method as in Sec.~\ref{sec:experiment}, except we project rendered scenes to a white background without floor plans. The results are presented in Table~\ref{tab:unconditional}.
We also note that in \cite{Tang24cvpr-DiffuScene}, DiffuScene used a 6.2m square to capture all synthesized layouts when generating the orthographic projection images for evaluation. Since the dining room and living room training data contains objects beyond this range, some of the predicted objects will be outside the projection range leading to incomplete result images for evaluation. In addition, the synthesized results were compared to the ground-truth training split. 
For completeness, we follow the evaluation setup in DiffuScene and present quantitative results in Table~\ref{tab:unconditional_additional} using the same predicted layouts as in Table~\ref{tab:unconditional}. The first row is directly copied from~\cite{Tang24cvpr-DiffuScene} demonstrating that we reproduced DiffuScene results close to the original publication values. The only exception is CA\%, for which DiffuScene does not release their classification network. Therefore, we adopt the AlexNet~\cite{Krizhevsky12nips-alexNet} based classifier from ATISS~\cite{Paschalidou21neurips-ATISS} to be consistent with Sec.~\ref{sec:experiment}.

Both Table~\ref{tab:unconditional} and ~\ref{tab:unconditional_additional} show that \algorithm achieves similar results, despite that its transformer-based denoising network is chosen for floor-conditioned scene synthesis.


\section{Additional Layout Examples}
We provide additional layout images generated by \algorithm in our floor-conditioned scene synthesis experiment. These layouts are generated in a sequence of random sampling of test floor plans after removing floor plan duplicates.
\clearpage

\begin{figure*}[!ht]
    \centering
    \begin{minipage}{.195\textwidth}
        \centering
        \includegraphics[width=0.95\linewidth]{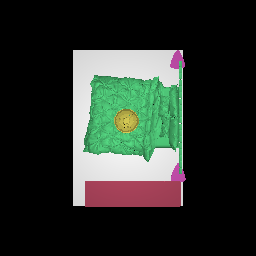}
    \end{minipage}%
    \hfill
    \begin{minipage}{0.195\textwidth}
        \centering
        \includegraphics[width=0.95\linewidth]{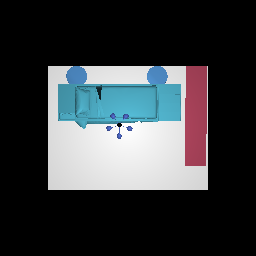}
    \end{minipage}
    \hfill
    \begin{minipage}{.195\textwidth}
        \centering
        \includegraphics[width=0.95\linewidth]{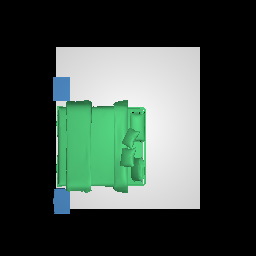}
    \end{minipage}%
    \hfill
    \begin{minipage}{.195\textwidth}
        \centering
        \includegraphics[width=0.95\linewidth]{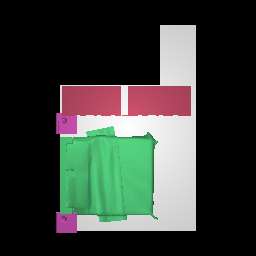}
    \end{minipage}%
    \hfill
    \begin{minipage}{.195\textwidth}
        \centering
        \includegraphics[width=0.95\linewidth]{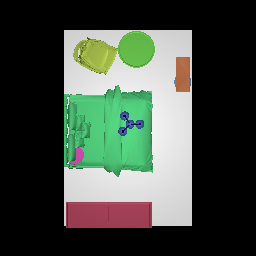}
    \end{minipage}%
    \\
    \begin{minipage}{.195\textwidth}
        \centering
        \includegraphics[width=0.95\linewidth]{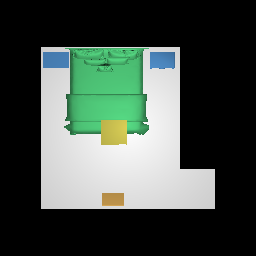}
    \end{minipage}%
    \hfill
    \begin{minipage}{.195\textwidth}
        \centering
        \includegraphics[width=0.95\linewidth]{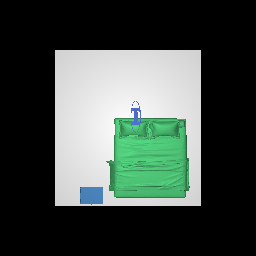}
    \end{minipage}%
    \hfill
    \begin{minipage}{.195\textwidth}
        \centering
        \includegraphics[width=0.95\linewidth]{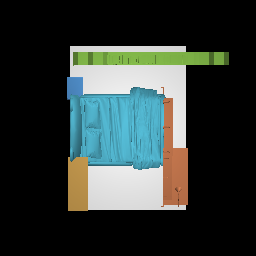}
    \end{minipage}%
    \hfill
    \begin{minipage}{.195\textwidth}
        \centering
        \includegraphics[width=0.95\linewidth]{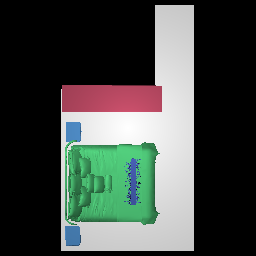}
    \end{minipage}%
    \hfill
    \begin{minipage}{.195\textwidth}
        \centering
        \includegraphics[width=0.95\linewidth]{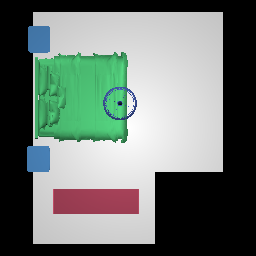}
    \end{minipage}%
    \vspace{0.2cm}

    \begin{minipage}{.195\textwidth}
        \centering
        \includegraphics[width=0.95\linewidth]{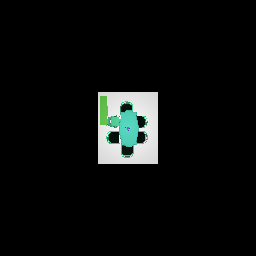}
    \end{minipage}%
    \hfill
    \begin{minipage}{0.195\textwidth}
        \centering
        \includegraphics[width=0.95\linewidth]{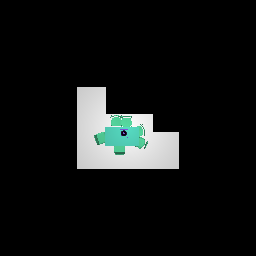}
    \end{minipage}
    \hfill
    \begin{minipage}{.195\textwidth}
        \centering
        \includegraphics[width=0.95\linewidth]{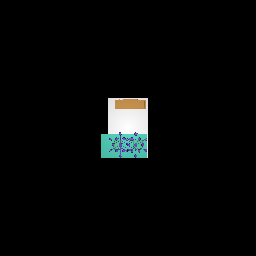}
    \end{minipage}%
    \hfill
    \begin{minipage}{.195\textwidth}
        \centering
        \includegraphics[width=0.95\linewidth]{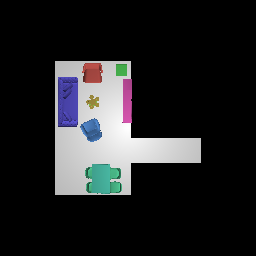}
    \end{minipage}%
    \hfill
    \begin{minipage}{.195\textwidth}
        \centering
        \includegraphics[width=0.95\linewidth]{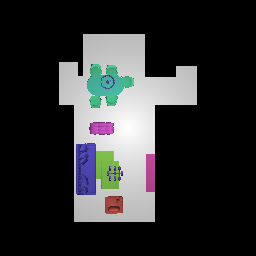}
    \end{minipage}%
    \\
    \begin{minipage}{.195\textwidth}
        \centering
        \includegraphics[width=0.95\linewidth]{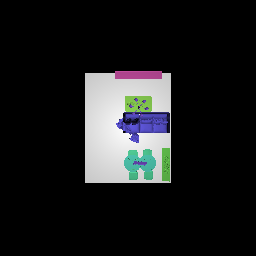}
    \end{minipage}%
    \hfill
    \begin{minipage}{.195\textwidth}
        \centering
        \includegraphics[width=0.95\linewidth]{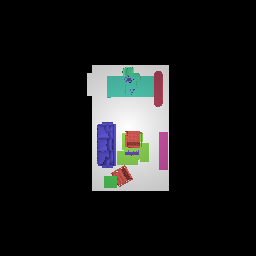}
    \end{minipage}%
    \hfill
    \begin{minipage}{.195\textwidth}
        \centering
        \includegraphics[width=0.95\linewidth]{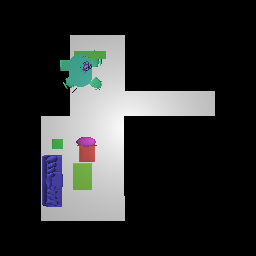}
    \end{minipage}%
    \hfill
    \begin{minipage}{.195\textwidth}
        \centering
        \includegraphics[width=0.95\linewidth]{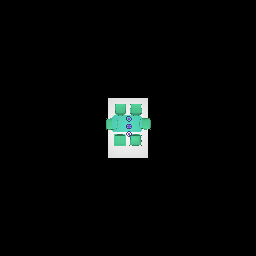}
    \end{minipage}%
    \hfill
    \begin{minipage}{.195\textwidth}
        \centering
        \includegraphics[width=0.95\linewidth]{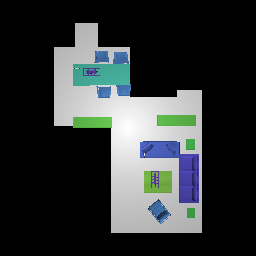}
    \end{minipage}%
    
    \vspace{0.2cm}

    \begin{minipage}{.195\textwidth}
        \centering
        \includegraphics[width=0.95\linewidth]{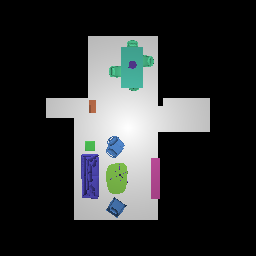}
    \end{minipage}%
    \hfill
    \begin{minipage}{0.195\textwidth}
        \centering
        \includegraphics[width=0.95\linewidth]{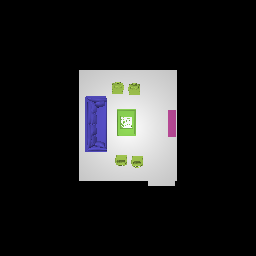}
    \end{minipage}
    \hfill
    \begin{minipage}{.195\textwidth}
        \centering
        \includegraphics[width=0.95\linewidth]{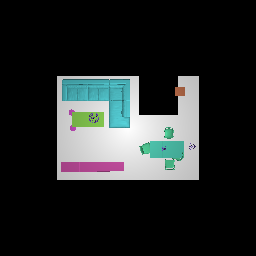}
    \end{minipage}%
    \hfill
    \begin{minipage}{.195\textwidth}
        \centering
        \includegraphics[width=0.95\linewidth]{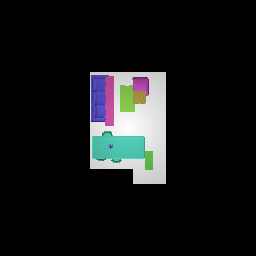}
    \end{minipage}%
    \hfill
    \begin{minipage}{.195\textwidth}
        \centering
        \includegraphics[width=0.95\linewidth]{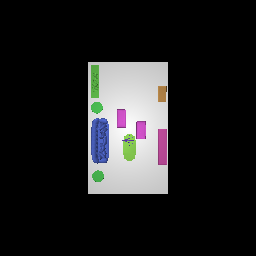}
    \end{minipage}%
    \\
    \begin{minipage}{.195\textwidth}
        \centering
        \includegraphics[width=0.95\linewidth]{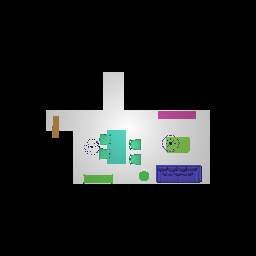}
    \end{minipage}%
    \hfill
    \begin{minipage}{.195\textwidth}
        \centering
        \includegraphics[width=0.95\linewidth]{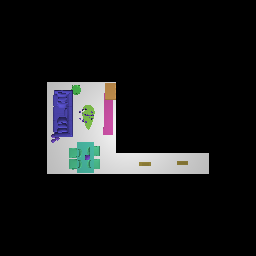}
    \end{minipage}%
    \hfill
    \begin{minipage}{.195\textwidth}
        \centering
        \includegraphics[width=0.95\linewidth]{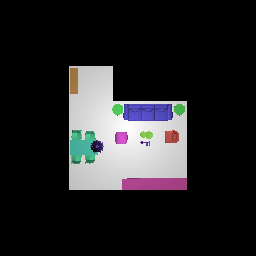}
    \end{minipage}%
    \hfill
    \begin{minipage}{.195\textwidth}
        \centering
        \includegraphics[width=0.95\linewidth]{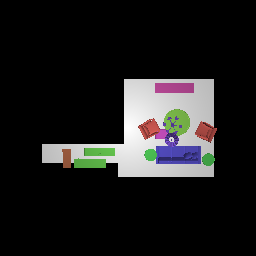}
    \end{minipage}%
    \hfill
    \begin{minipage}{.195\textwidth}
        \centering
        \includegraphics[width=0.95\linewidth]{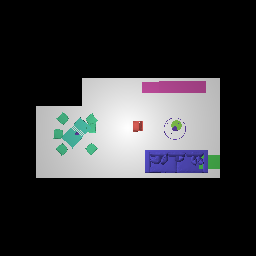}
    \end{minipage}%
\end{figure*}

\end{document}